\documentclass[journal]{IEEEtran}
\usepackage{graphicx}
\usepackage{subcaption}
\usepackage[table,xcdraw]{xcolor}
\usepackage{multirow}
\usepackage{adjustbox}
\usepackage{subfig}
\usepackage{booktabs}
\usepackage{cite}
\usepackage{bbding}

\definecolor{others}{rgb}{0, 0, 0}
\definecolor{barrier}{rgb}{1, 0.47058824, 0.19607843}
\definecolor{bicycle}{rgb}{1, 0.75294118, 0.79607843}
\definecolor{bus}{rgb}{1, 1, 0.0}
\definecolor{car}{rgb}{0.0, 0.58823529, 0.96078431}
\definecolor{construction}{rgb}{0, 1, 1}
\definecolor{motorcycle}{rgb}{1, 0.49803922, 0}
\definecolor{pedestrian}{rgb}{1, 0, 0}
\definecolor{cone}{rgb}{1, 0.94117647, 0.58823529}
\definecolor{trailer}{rgb}{0.52941176, 0.23529412, 0}
\definecolor{truck}{rgb}{0.62745098, 0.1254902, 0.94117647}

\definecolor{driveable}{rgb}{1, 0, 1}
\definecolor{flat}{rgb}{0.54509804,0.5372549,0.5372549}
\definecolor{sidewalk}{rgb}{0.29411765,0,0.29411765}
\definecolor{terrain}{rgb}{0.58823529,0.94117647,0.31372549}
\definecolor{manmade}{rgb}{0.90196078,0.90196078,0.98039216}
\definecolor{vegetation}{rgb}{0,0.68627451,0}
\definecolor{ego_vehicle}{rgb}{0,0,0}
\ifCLASSINFOpdf
\else
\fi
\begin{document}
%
\title{OccFusion: Multi-Sensor Fusion Framework for 3D Semantic Occupancy Prediction}
%
%
%

\author{Zhenxing~Ming, Julie~Stephany~Berrio, Mao~Shan, and Stewart~Worrall
\thanks{This work has been supported by the Australian Centre for Robotics (ACFR). The authors are with the ACFR at the University of Sydney (NSW, Australia). E-mails: zmin2675@uni.sydney.edu.au, \{j.berrio, m.shan, s.worrall\}@acfr.usyd.edu.au}%
}

\maketitle

\begin{abstract}
 A comprehensive understanding of 3D scenes is crucial in autonomous vehicles (AVs), and recent models for 3D semantic occupancy prediction have successfully addressed the challenge of describing real-world objects with varied shapes and classes. However, existing methods for 3D occupancy prediction heavily rely on surround-view camera images, making them susceptible to changes in lighting and weather conditions. This paper introduces OccFusion, a novel sensor fusion framework for predicting 3D occupancy. By integrating features from additional sensors, such as lidar and surround view radars, our framework enhances the accuracy and robustness of occupancy prediction, resulting in top-tier performance on the nuScenes benchmark. Furthermore, extensive experiments conducted on the nuScenes and semanticKITTI dataset, including challenging night and rainy scenarios, confirm the superior performance of our sensor fusion strategy across various perception ranges. The code for this framework will be made available at https://github.com/DanielMing123/OccFusion.
\end{abstract}


%
\IEEEpeerreviewmaketitle

\section{Introduction}
\begin{figure}[t]
\centering
{\includegraphics[width=\columnwidth]{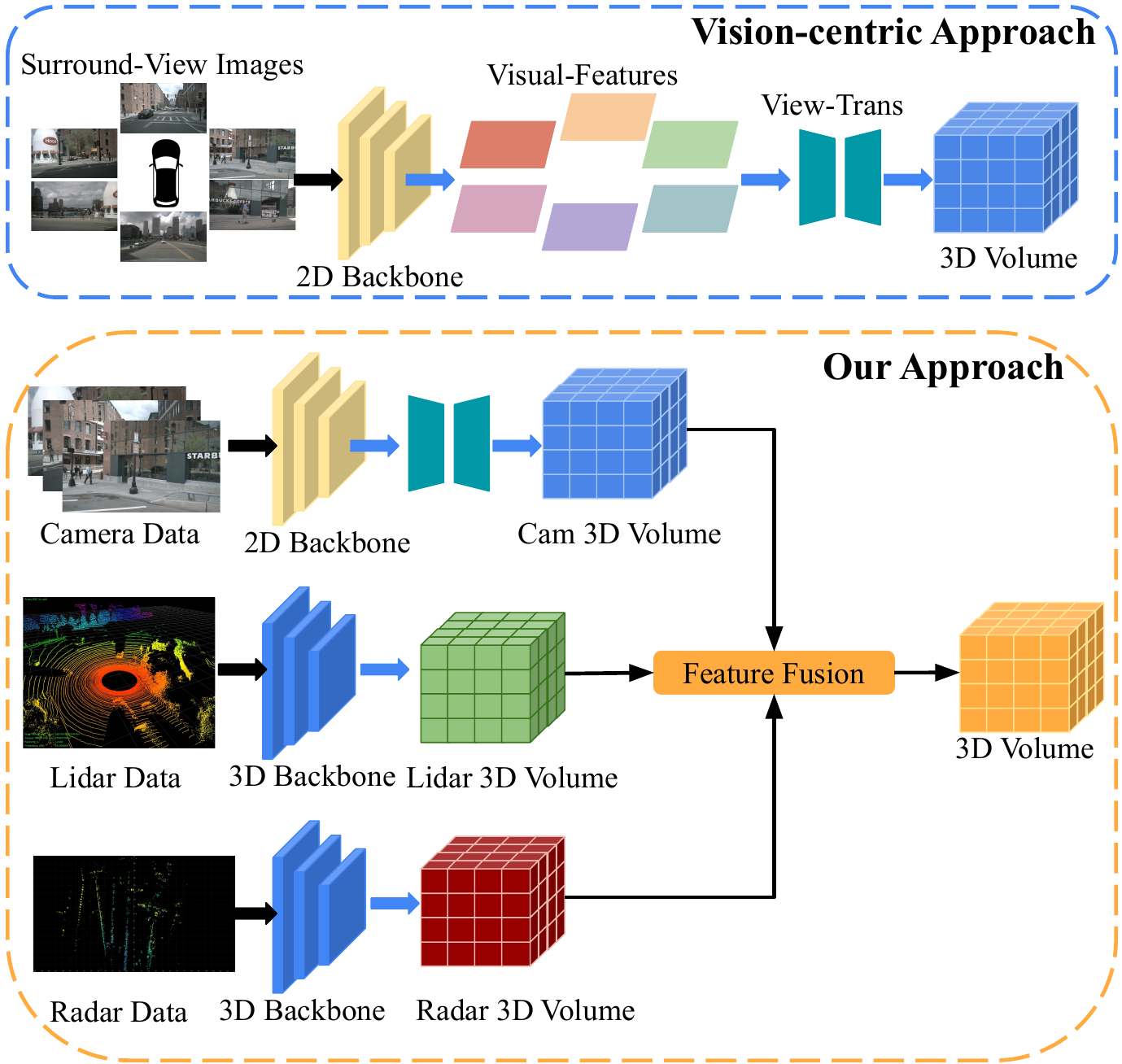}}
\caption{Pipeline for two approaches: purely vision-centric approach (top) and multi-sensor fusion approach (bottom). We conduct 3D semantic occupancy prediction by doing feature fusion with respect to three modality feature volumes.}\label{intro}
\end{figure}

\IEEEPARstart{U}{nderstanding} and modelling the three-dimensional (3D) world is essential for autonomous vehicles (AVs) to navigate safely, preventing collisions and facilitating local planning. As technology advances, the introduction of 3D occupancy representation has successfully addressed the limitations of traditional 3D object detection networks, particularly for detecting irregular objects and space-occupied status prediction. This advancement further enhances the capability of AVs to model the 3D world. However, the currently proposed 3D occupancy prediction models \cite{monoscene,bevdet,bevformer,bevstereo,tpvformer,occ3d,surroundocc,radocc,renderocc,bevdet4d,panoocc,fbocc,octreeocc,inversematrixvt3d}, primarily focus on vision-based approaches (Figure \ref{intro}, upper). While surround-view cameras are cost-effective, their perception capabilities are highly susceptible to poor weather conditions like rain or fog, and illumination conditions such as those experienced at night. These factors cause the model to perform inconsistently in these scenarios, posing potential safety risks.

Besides the surround-view cameras, AVs are often equipped with lidars and surround-view millimetre wave radars. Lidar excels at capturing the geometric shape of objects and accurately measuring depth. Moreover, it is robust to illumination changes and performs reliably under various weather conditions, except for heavy rain and fog. In contrast, surround-view millimetre wave radars are cost-effective and exceptionally robust against weather conditions and changes in illumination. However, they yield only sparse features, which are often noisy. 
Each sensor has its advantages and disadvantages.  Combining information from all three sensors (Figure \ref{intro}, bottom) can potentially enhance AVs' 3D occupancy prediction model to achieve superior accuracy when modelling the 3D world. Additionally, this integration improves the system's robustness in the face of varying lighting and weather conditions.

To predict 3D occupancy accurately and efficiently, we propose a framework called OccFusion. This framework merged features from surround-view cameras, surround radars, and 360-degree lidar through dynamic fusion 3D/2D modules. There are three sensor fusion strategies: Camera + Radar, Camera + Lidar, and Camera + Lidar + Radar. To evaluate the accuracy of our sensor fusion strategies, we conduct experiments on the nuScenes dataset \cite{nuscenes}, utilizing ground truth labels provided by the SurroundOcc \cite{surroundocc} and Occ3D \cite{occ3d}. Additionally, we manually select rainy and nighttime scenarios from the nuScenes validation set to create two challenging subsets and examine the performance of different sensor fusion strategies in these scenarios. Finally, we test how the performance of different sensor fusion strategies varies across different perception ranges under different scenarios.

The main contributions of this paper are summarized as below. 
\begin{itemize}
    \item A multi-sensor fusion framework is proposed to integrate camera, lidar, and radar information for the 3D semantic occupancy prediction task.
    \item We compared our approach with other state-of-the-art (SOTA) algorithms in the 3D semantic occupancy prediction task to prove the advantages of multi-sensor fusion.
    \item We conducted thorough ablation studies to assess the performance gains of different sensor combinations under challenging lighting and weather conditions, such as night and rainy scenarios.
    \item We conducted a comprehensive study to analyze the influence of perception range factors on the performance of our framework in 3D semantic occupancy prediction tasks, considering various sensor combinations and challenging scenarios.
\end{itemize}

The remainder of this paper is structured as follows: Section \ref{literature} provides an overview of related research and identifies the key differences between this study and previous publications. Section \ref{model} outlines the general framework of OccFusion and offers a detailed explanation of the implementation of each module. Section \ref{simulation} presents the findings of our experiments. Finally, Section \ref{conclusion} provides the conclusion of our work.

\section{Related Work}\label{literature}
This section presents the most recent research findings on various sensor fusion algorithms used for environmental perception in autonomous driving contexts.

\subsection{Camera-only-based environment perception}
In recent years, surround-view camera-based environment perception algorithms have received significant attention in the AVs domain due to their cost-effectiveness and versatility. The bird's-eye-view (BEV) feature-based algorithms \cite{LSS,bevdet,bevdepth,bevstereo,bevformer,bevformerv2,beverse} successfully merged all surround-view camera information to conduct 3D object detection tasks. By lifting the BEV feature into 3D feature volume, algorithms \cite{bevdet,bevdet4d} are capable of doing 3D semantic occupancy tasks. Two main approaches to view transformation in these algorithms are the classic Lift-Splat-Shoot (LSS) and transformer-based approaches. The LSS-based approach \cite{fbocc,occformer} relies on depth estimation to generate a pseudo-3D point cloud, followed by voxel-pooling to create the final 3D feature volume. On the other hand, the transformer-based approach \cite{tpvformer,panoocc,occ3d,octreeocc,radocc,renderocc,surroundocc} uses sampling points to aggregate visual features from feature maps and places these features directly at specific 3D positions in the world to form the final 3D feature volume. Both approaches explicitly estimate the depth or implicitly encode depth information in visual features. Nonetheless, it is well-known that monocular cameras are inadequate for accurate depth estimation. While they can capture the relative depth position of an object, they cannot provide precise depth information. Hence, a more reliable reference for depth information is required. This could involve incorporating lidar information into the model to enhance depth estimation or using lidar information to supervise depth estimation, as in the BEVDet series approaches.

\subsection{Lidar-only-based environment perception}
Lidar-only-based algorithms \cite{adabins,AF2S3Net,amvnet,polarnet,polarstream,JS3C-Net,minet,SPVNAS,cylinder3d,DRINet++,LidarMultiNet} for environment perception have shown promising performance in various perception tasks. Leveraging its capability for accurate depth estimation, lidar excels in capturing the geometric shape and 3D location of objects. By converting the 3D point cloud into Euclidean feature spaces, such as 3D voxel grids \cite{voxelnet} or feature pillars \cite{pointpillars}, lidar-based methods can achieve highly precise 3D object detection results. In recent years, researchers have extended lidar's 3D point cloud features into 3D semantic occupancy prediction tasks \cite{4d-occ,occ4cast}. However, the density of the lidar-generated 3D point cloud strongly influences the final perception performance of the model, and its lack of semantic information results in inaccurate object class recognition. Hence, the auxiliary information is needed to provide comprehensive semantic information guidance, which leads to our work fusing the lidar data with camera data to enhance the performance of 3D semantic occupancy prediction.

\subsection{Camera-Lidar fusion-based environment perception}
Due to the inherent advantages and disadvantages of individual sensors, recent research has focused on sensor fusion techniques \cite{f-pointnet,iPod,SIFRNet,pointpainting,MVX-Net,ContFuse} to overcome these limitations and enhance the overall environment perception capability of the models. The representative BEVFusion \cite{bevfusion1,bevfusion2} algorithms fuse lidar and surround-view cameras by encoding the features of each modality into BEV features and performing feature fusion. This approach addresses the reflection issues lidar encounters in rainy and foggy scenarios, often resulting in false and missed detections. It also resolves monocular cameras' poor depth estimation problem, enabling the model to generate relatively accurate detection results at longer distances. The SparseFusion \cite{sparsefusion} further refines the inner structure of the feature fusion module, greatly improving the model's inference speed. However, currently, 
most existing algorithms primarily serve for 3D object detection. Therefore, there is a compelling need for extensive research on camera-lidar fusion techniques for 3D semantic occupancy prediction.

\subsection{Camera-Radar fusion-based environment perception}
Various studies \cite{guizilini2022full,EZFusion,seeing} have been conducted on the fusion of cameras and radar for environmental perception due to radar's cost-effectiveness and ability to detect distant objects. For instance, the work in \cite{vehicle} demonstrates that information about velocity obtained from radar sensors can enhance detection performance. Additionally, a study by \cite{centerfusion} suggests that integrating radar features with visual features can result in a performance gain of approximately 12\% under the nuScene Detection Score (NDS) metric. Furthermore, another study in \cite{radar} found that radar sensor readings exhibit robustness in noisy conditions, and integrating radar information can improve model performance in challenging scenarios. Though various algorithms have been developed to achieve camera-radar fusion, most focus on 3D object detection, object tracking, and object future trajectory prediction tasks. No camera-radar fusion algorithm is available specifically for 3D semantic occupancy prediction tasks. Moreover, this task requires dense features, whereas radar provides sparse features. Consequently, we investigate the impact on model performance when merging these sparse radar features with camera and lidar data. To the best of our knowledge, our study is the first to examine the influence of fused radar information on a 3D semantic occupancy prediction task. 

\subsection{Camera-Lidar-Radar fusion-based environment perception}
Due to the complementary property of multi-sensor fusion, people in this domain also investigate the camera-lidar-radar fusion strategy and its performance in environment perception. In CLR-BNN \cite{clrbnn}, the authors employ a Bayesian neural network for camera-lidar-radar sensor fusion, yielding improved 2D multi-object detection results in terms of mAP. In Futr3D \cite{futr3d}, the sensor fusion is further explored through the incorporation of transformers, using sparse 3D points as queries to aggregate features from the three sensors for the 3D object detection task, the query-form significantly improved feature interactions and aggregation efficiency across three sensors. In SimpleBEV \cite{simplebev}, authors process data from all three sensors into BEV features, fusing these features based on three BEV representations to perform the 3D object detection task; in their research, they found radar data provide a substantial boost to performance. Previous research has extensively examined the characteristics of three sensor fusion methods for environmental perception. However, these works have primarily concentrated on 2D or 3D multi-object detection tasks, neglecting investigations into the 3D semantic occupancy prediction task. Therefore, there is a need to investigate the performance of the camera-lidar-radar sensor fusion strategy in the context of 3D semantic occupancy prediction tasks.

\section{OccFusion}\label{model}
\begin{figure*}[h]
\vspace{2mm}\
     \centering
     \begin{subfigure}[]{\textwidth}
         \centering
         \includegraphics[width=\textwidth]{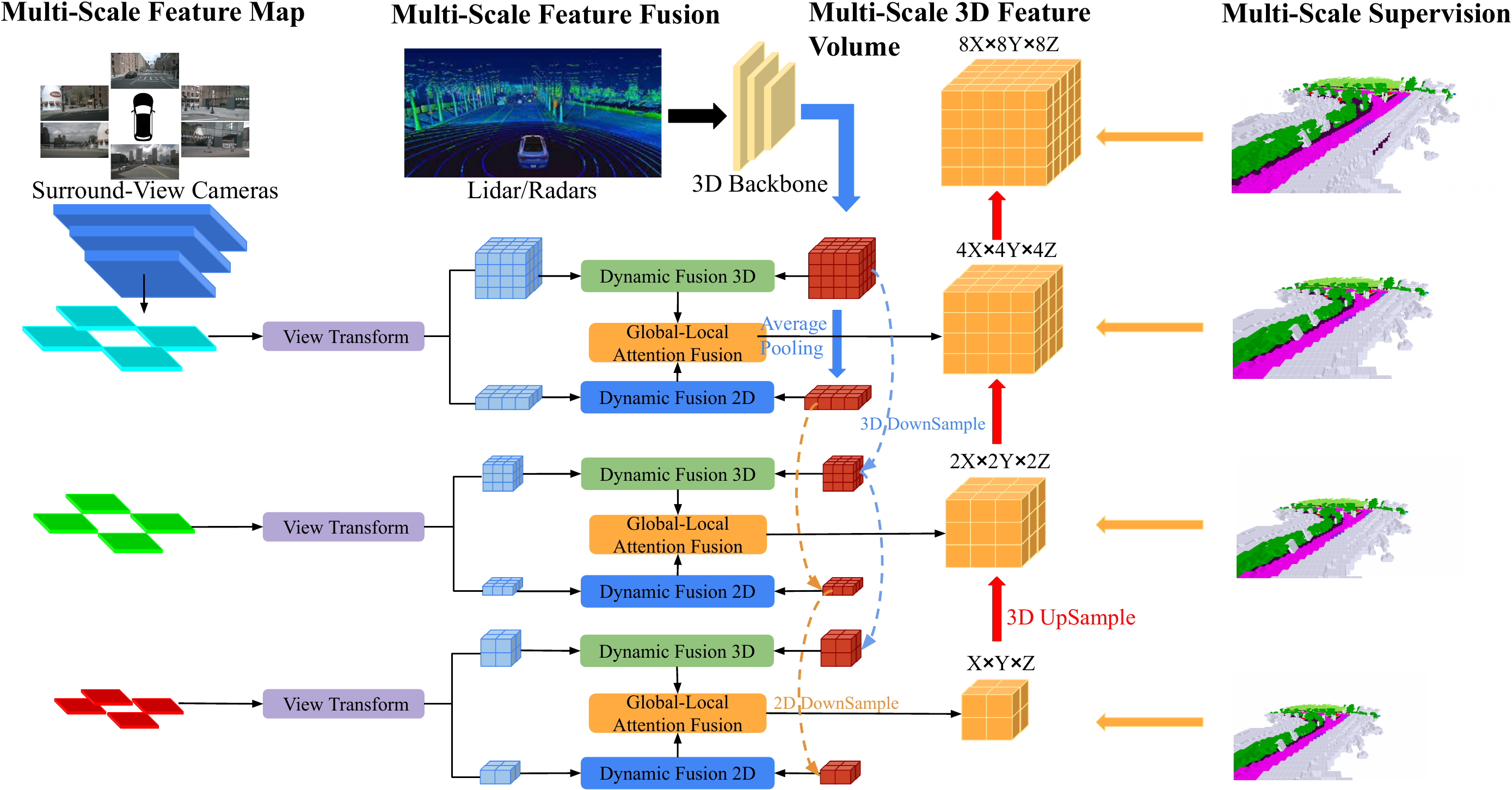}
     \end{subfigure}
        \caption{\small \textbf{Overall architecture of OccFusion.} Firstly, the surround-view images were inputted into the 2D backbone to extract multiple-scale features. Subsequently, each scale's view transformation is conducted to obtain each level's global BEV feature and the local 3D feature volume. The 3D point cloud generated by the lidar and surround-view radars is also inputted into the 3D backbone to generate multi-scale local 3D feature volumes and global BEV features. The dynamic fusion 3D/2D modules at each level fuse features from the cameras and lidar/radar. Following this, each level's merged global BEV feature and local 3D feature volume are fed into the global-local attention fusion to generate the final 3D volume at each scale. Finally, the 3D volume at each level is upsampled, and the skip connection is performed while adopting a multi-scale supervision mechanism.}
        \label{OccFusion}
\end{figure*}

\subsection{Problem Statement}
This paper aims to generate a dense 3D semantic occupancy grid of the surrounding scene by integrating information from surround-view cameras $Cam=\left \{Cam^{1},Cam^{2},\cdots,Cam^{N} \right \}$, surround-view radars $Rad=\left \{Rad^{1},Rad^{2},\cdots,Rad^{N} \right \}$, and lidar $Lid$. Thus, the problem can be formulated as:
\begin{equation}
    Occ = F(Cam^{1},\dots,Cam^{N},Rad^{1},\cdots,Rad^{N},Lid)
\end{equation}
where $F$ represents the fusion framework integrating multi-sensor information for 3D occupancy prediction. The final 3D occupancy prediction result is represented by $Occ \in R^{X \times Y \times Z}$, where each grid is assigned a semantic property ranging from 0 to 17. In our case, a class value of 0 corresponds to an empty grid.

\subsection{Overall Architecture}
Figure \ref{OccFusion} exhibits the overall architecture of our proposed framework. Initially, given surround-view images, dense 3D point clouds $P^{Dense}$ from lidar and sparse 3D point clouds $P^{Sparse}$ from surround-view radars, we apply a 2D backbone (e.g. ResNet101-DCN) to extract total $L$ scale features $M=\left \{ \left \{ M_{n}^{l}  \right \}_{n=1}^{N}\in R^{C_{l} \times H_{l} \times W_{l}} \right \} _{l=1}^{L}$ from images, followed by view transformation to obtain the global BEV feature $F_{global}^{Cam\_l}\in  R ^{C_{l}\times X_{l} \times Y_{l}} $ and local 3D feature volume $F_{local}^{Cam\_l}\in  R ^{C_{l}\times X_{l} \times Y_{l} \times Z_{l}}$ at each scale. Meanwhile, a 3D backbone (e.g. VoxelNet) is also applied to dense and sparse 3D point clouds to generate multi-scale global BEV features $F_{global}^{Rad\_l}\in  R ^{C_{l}\times X_{l} \times Y_{l}} $, $F_{global}^{Lid\_l}\in  R ^{C_{l}\times X_{l} \times Y_{l}} $ and local 3D feature volumes $F_{local}^{Rad\_l}\in  R ^{C_{l}\times X_{l} \times Y_{l} \times Z_{l}}$, $F_{local}^{Lid\_l}\in  R ^{C_{l}\times X_{l} \times Y_{l} \times Z_{l}}$, respectively. Following that, the $F_{global}^{Cam\_l}$, $F_{global}^{Rad\_l}$ and $F_{global}^{Lid\_l}$ at each level are fed into the dynamic fusion 2D module to obtain the merged global BEV feature $F_{global}^{Merged\_l}$. Simultaneously, the $F_{local}^{Cam\_l}$, $F_{local}^{Rad\_l}$ and $F_{local}^{Lid\_l}$ at each level are also fed into the dynamic fusion 3D module to obtain the merged local 3D feature volume $F_{local}^{Merged\_l}$. Subsequently, the global-local attention fusion module proposed in \cite{inversematrixvt3d} is used at each level to merge further $F_{local}^{Merged\_l}$ and $F_{global}^{Merged\_l}$, resulting in the final 3D volume at each level. Moreover, a skip-connection structure is implemented between each level to refine the features in a coarse-to-fine manner, and multi-scale supervision is applied to improve the model's performance.

\subsection{Surround-View Images Feature Extraction}
Given surround-view images, we initially employ ResNet101-DCN \cite{resnet}  as our 2D backbone and feature pyramid network (FPN) \cite{fpn} as the neck to extract multi-scale feature maps. The resulting feature maps have resolutions that are $\frac{1}{8}$, $\frac{1}{16}$, and $\frac{1}{32}$ of the input image resolution, respectively. Subsequently, a view transformation is leveraged, yielding multi-scale global BEV features and local 3D feature volumes. The global and local features with smaller resolutions contain valuable semantic information, aiding the model in predicting the semantic class of each voxel grid. Conversely, those features with larger resolutions provide rich spatial information, enabling the model to determine whether the current voxel grid is occupied or unoccupied.

\subsection{Lidar Dense 3D Point Cloud Feature Extraction}
In this paper, we adopt VoxelNet \cite{voxelnet} as our 3D backbone for feature extraction of the 3D point cloud. We begin by voxelizing the 3D point cloud to generate the voxel grid and its associated coordinates. In each voxel grid containing 3D points, 35 points are randomly selected. In each voxel grid containing 3D points, 35 points are randomly selected. If a voxel grid has fewer than 35 points, zero padding is applied to reach 35 points. Each point, denoted as $p_i^{Lid}$, includes an initial feature vector $p_i^{Lid} =[x_i, y_i, z_i, \gamma_i]$ that represents the point's 3D position and reflectance rate. Subsequently, we calculate the centre position of these 3D points within the same voxel grid and augment each point with a relative offset with respect to the centre position. This augmentation results in a new feature vector $p_i^{Lid} =[x_i, y_i, z_i, \gamma_i, x_i-\bar{x}, y_i-\bar{y}, z_i-\bar{z}]$. Following this, only the voxel grid with 3D points is input to the 3D backbone to refine the features further, producing the final local 3D feature volume. Additionally, we employ average pooling along the Z-axis of the local 3D feature volume for the global BEV feature to obtain the flattened global BEV feature. The 3D backbone outputs the highest resolution of the global and local features, while the lower resolution features are obtained through 3D/2D downsampling operations.

\subsection{Radar Sparse 3D Point Cloud Feature Extraction}
The Radar 3D point contains richer information compared to the lidar 3D point. Each Radar 3D point has an initial feature vector $p_i^{Rad} =[x_i,y_i,z_i,V_{xi},V_{yi}]$, where $V_{xi}$ denote the velocity along the X-axis, and $V_{yi}$ denote the velocity along the Y-axis. Like the lidar point cloud processes, we begin by voxelizing the Radar 3D point cloud and obtaining the voxel grids and their associate coordinates. For the voxel grid with points, we calculate the mean value among the 3D points and augment each point with a relative offset with respect to the mean value. This augmentation results in a new feature vector $p_i^{Rad} =[x_i,y_i,z_i,V_{xi},V_{yi},x_i-\bar{x}, y_i-\bar{y}, z_i-\bar{z},V_{xi}-\bar{V_x},V_{yi}-\bar{V_{y}}]$. We subsequently input the non-empty voxel grid into the 3D backbone to obtain the local 3D feature volume and then apply average pooling to obtain the global BEV feature.

\subsection{Dynamic Fusion 3D/2D}
\begin{figure}[htbp]
\centering
{\includegraphics[width=\columnwidth]{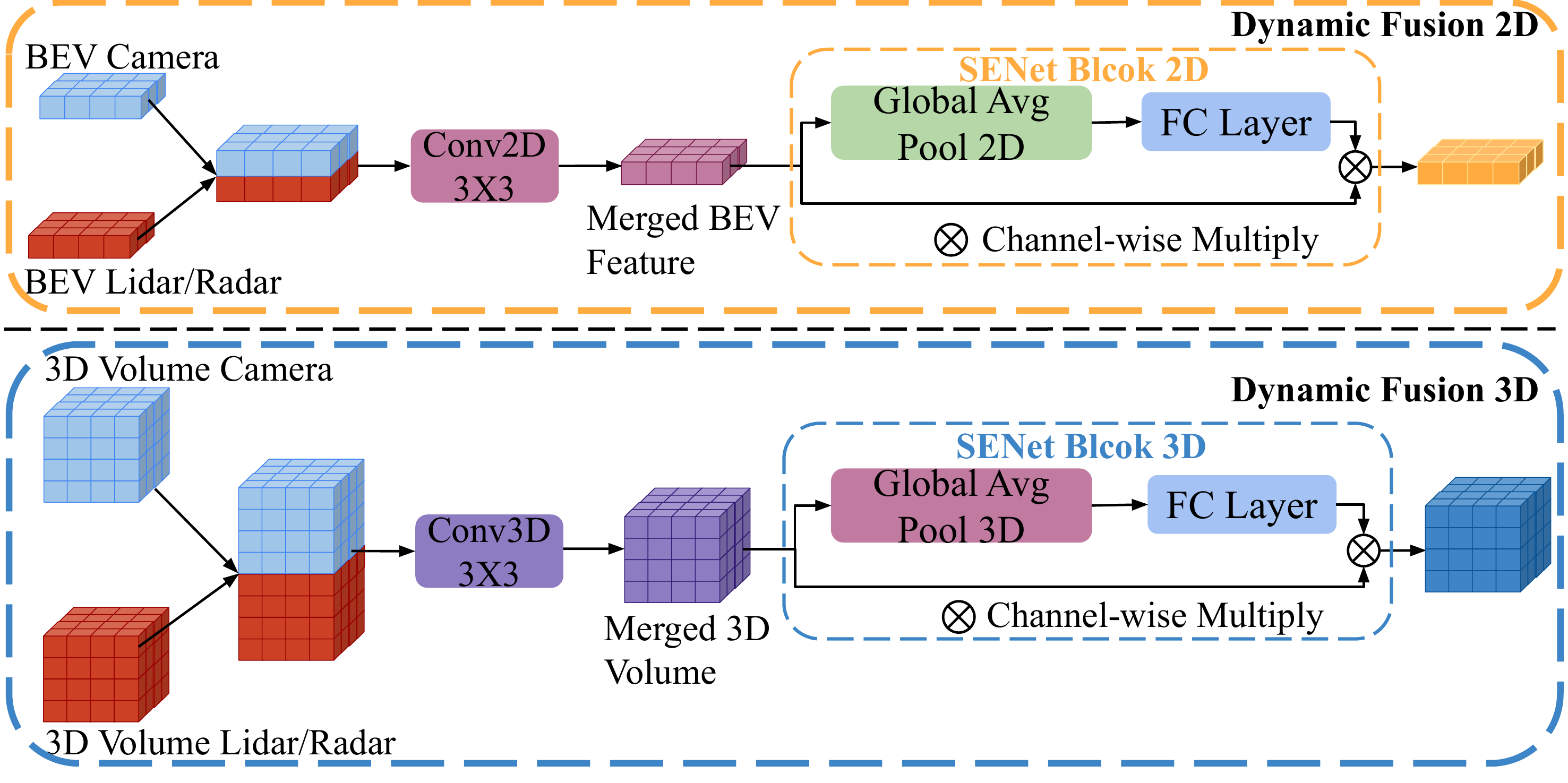}}
\caption{Dynamic Fusion 3D/2D Modules. The upper diagram exhibits the process details of the dynamic fusion 2D module, and the bottom diagram shows the process details of the dynamic fusion 3D module.}\label{dynamic}
\end{figure}

Drawing inspiration from BEVFusion \cite{bevfusion1, bevfusion2} and SENet \cite{senet}, this research merges two BEV features and two 3D feature volumes by concatenating their feature channels. Subsequently, a Conv3D/2D layer is applied to reduce the feature channel dimension, facilitating the merging of valuable features from diverse modalities while filtering out noisy features. This process is followed by a 3D/2D SENet block, where the merged features are inputted into the squeeze module to determine the importance of each feature channel. The excitation module then implements an excitation procedure by multiplying the merged features with the squeeze feature, enabling critical features to dominate. Details of the Dynamic Fusion 2D module are in the upper section of Figure \ref{dynamic}. In contrast, specifics of the Dynamic Fusion 3D module are depicted in the bottom section of Figure \ref{dynamic}.

\section{Experimental Result}\label{simulation}
\subsection{Implementation Details}
The OccFusion utilizes ResNet101-DCN \cite{residual, deformable} as the 2D backbone, with pre-trained weights provided by FCOS3D \cite{fcos3d}, to extract image features. The backbone's feature maps from stages 1, 2, and 3 are then fed into FPN \cite{fpn}, resulting in three levels of multi-scale image features. The network architecture consists of four levels ($L=4$), with no skip connection applied to the highest level. Our framework is adaptable to any view transformation approach. In this paper, we choose the view transformation method proposed in InverseMatrixVT3D \cite{inversematrixvt3d} to aggregate visual features. It is worth mentioning that when the framework does not merge lidar and radar information, it is the same as the InverseMatrixVT3D algorithm. Thus, OccFusion (C) has the same performance as InverseMatrixVT3D. Our framework uses 10 lidar sweeps, and 5 surround-view radar sweeps for each data sample. To extract 3D feature volumes, the VoxelNet \cite{voxelnet} is used as the framework's 3D backbone, which processes the dense 3D point cloud from lidar and the sparse 3D point cloud from surround-view radars. The AdamW optimizer with an initial learning rate of 5e-5 and weight decay of 0.01 is employed for optimisation. The learning rate is decayed using a multi-step scheduler. The model is trained on eight A10 GPUs, each with 24GB of memory for two days.

\subsection{Loss Function}
The framework is trained to utilize focal loss \cite{focal}, Lovasz-Softmax loss \cite{lovasz}, and scene-class affinity loss \cite{monoscene}. Considering the significance of high-resolution 3D volumes compared to lower-resolution ones, a decayed loss weight $w=\frac{1}{2^l}$ is applied for supervision at the $l$-th level. The ultimate loss formulation is as follows:
\begin{equation}
    Loss = \sum_{l=0}^{3} \frac{1}{2^{l} } \times (L_{focal}^{l} + L_{lovasz}^{l} + L_{scal\_geo}^{l} + L_{scal\_sem}^{l})
\end{equation}
where $l$ represent $l$-th level within the framework.

\subsection{Dataset}

\begin{figure*}[t]
     \centering
     \begin{subfigure}[]{0.32\textwidth}
         \centering
         \includegraphics[width=\textwidth]{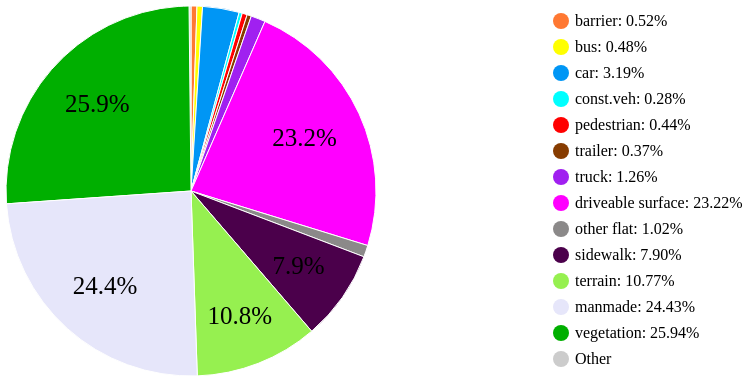}
         \caption{}
         \label{val}
     \end{subfigure}
     \begin{subfigure}[]{0.32\textwidth}
         \centering
         \includegraphics[width=\textwidth]{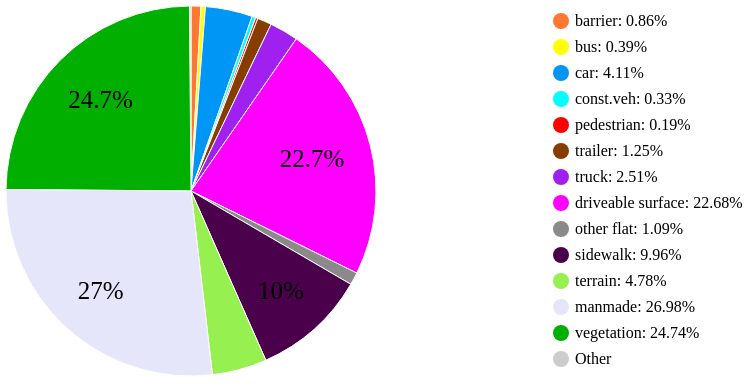}
         \caption{}
         \label{val_rainy}
     \end{subfigure}
     \begin{subfigure}[]{0.32\textwidth}
         \centering
         \includegraphics[width=\textwidth]{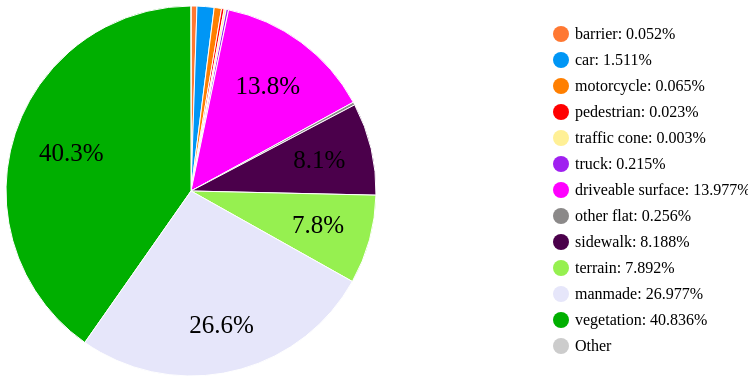}
         \caption{}
         \label{val_night}
     \end{subfigure}
        \caption{\small Class distribution for three validation sets. (a) whole validation set class distribution, (b) rainy scenario subset class distribution, and (c) night scenario subset class distribution.}
        \label{cls_dist}
\end{figure*}
Our 3D semantic occupancy prediction and 3D scene completion experiments were carried out using the nuScenes dataset. The ground truth labels utilized in these experiments were sourced from the works of SurroundOcc \cite{surroundocc} and Occ3D \cite{occ3d}. The SurroundOcc's label spans a range of -50 m to 50 m for the X and Y directions and -5 m to 3 m for the Z direction. This range suits our model's ablation study on the perception range factor. On the other hand, the Occ3D provide a ground truth label for a relatively minor perceptual range, which spans a range of -40 m to 40 m for the X and Y directions and -1 m to 5.4 m for the Z direction. Furthermore, since Occ3D's label was designed for purely vision-centric algorithms, a visibility mask is provided for each voxel grid, and evaluation only considers visible voxels. As the test set labels were unavailable, we trained our model on the training set and evaluated its performance on the validation set. Additionally, we chose particular frames from the nuScenes validation set, using ground truth labels from SurroundOcc's work to establish subsets corresponding to rainy and night scenarios. The distribution of classes in the validation set, rainy scenario subset, and night scenario subset is shown in Figure \ref{cls_dist}. The class with a sample number equal to zero is not listed in each set.

To further validate the efficacy of our methodology, we conducted a semantic scene completion experiment on the SemanticKITTI dataset utilizing data from the left RGB camera and lidar. SemanticKITTI provides annotated outdoor lidar scans classified into 21 semantic classes. The input image has a resolution of $1241\times376$, and the ground truth is voxelized into a $256\times256\times32$ grid with a voxel size of 0.2m. Given this dataset's absence of radar sensors, we assessed our OccFusion (C+L) model on the validation set.

\subsection{Performance Evaluate Metrics}
In assessing the effectiveness of various SOTA algorithms for 3D semantic occupancy prediction and contrasting them with our approach, we utilize Intersection over Union (IoU) to evaluate each semantic class. Moreover, we adopt IoU for occupied voxels, disregarding their semantic class for the scene completion task evaluation. Additionally, the mean IoU across all semantic classes (mIoU) serves as a comprehensive evaluation metric:
\begin{equation}
    IoU=\frac{TP}{TP+FP+FN} 
\end{equation}
and
\begin{equation}
    mIoU=\frac{1}{Cls}\sum_{i=1}^{Cls}  \frac{TP_{i}}{TP_{i}+FP_{i}+FN_{i}} 
\end{equation}
where $TP$, $FP$, and $FN$ represent the counts of true positives, false positives, and false negatives in our predictions, respectively, while $Cls$ denotes the total class number.

\subsection{Model Performance Analysis}
We assess different sensor fusion strategies using our proposed framework on the nuScenes validation set and present the final benchmark results in Table \ref{occ} and Table \ref{occ_3d}. In Table \ref{occ}, compared to purely vision-centric approaches, including sparse 3D point cloud information from the surround-view radars leads to a noteworthy around 2\% improvement in performance. Moreover, adding dense 3D point cloud information from the lidar further enhances the performance to approximately 27\% mIoU. These experimental results confirm the efficacy of utilizing multi-sensor fusion to substantially improve the performance of the 3D semantic occupancy prediction task. However, in Table \ref{occ_3d}, we observed a non-trivial performance degradation caused by merging radar information. This may contribute to the nature of radar, which is good at measuring objects that are far away, but the perception range in Occ3D's label and the visibility mask limit its strength. 

To further evaluate the effectiveness of our proposed framework, we conduct comparative experiments for the Semantic Scene Completion task on the SemanticKITTI dataset. The benchmark result is presented in Table \ref{kitti}. Integrating camera and lidar data in our framework yields highly competitive performance compared to vision-centric and lidar-centric algorithms.
\begin{table*}[htpb]
  \centering
  \begin{adjustbox}{width=\textwidth}
  \begin{tabular}{c|c|c|cc|cccccccccccccccc}
    \toprule
    Method & Backbone & Input Modality & IoU & mIoU & \rotatebox{90}{\textcolor{barrier}{$\bullet$} barrier} & \rotatebox{90}{\textcolor{bicycle}{$\bullet$} bicycle} & \rotatebox{90}{\textcolor{bus}{$\bullet$} bus} & \rotatebox{90}{\textcolor{car}{$\bullet$} car} & \rotatebox{90}{\textcolor{construction}{$\bullet$} const. veh.} & \rotatebox{90}{\textcolor{motorcycle}{$\bullet$} motorcycle} & \rotatebox{90}{\textcolor{pedestrian}{$\bullet$} pedestrian} & \rotatebox{90}{\textcolor{cone}{$\bullet$} traffic cone} & \rotatebox{90}{\textcolor{trailer}{$\bullet$} trailer} & \rotatebox{90}{\textcolor{truck}{$\bullet$} truck} & \rotatebox{90}{\textcolor{driveable}{$\bullet$} drive. surf.} & \rotatebox{90}{\textcolor{flat}{$\bullet$} other flat} & \rotatebox{90}{\textcolor{sidewalk}{$\bullet$} sidewalk} & \rotatebox{90}{\textcolor{terrain}{$\bullet$} terrain} & \rotatebox{90}{\textcolor{manmade}{$\bullet$} manmade} & \rotatebox{90}{\textcolor{vegetation}{$\bullet$} vegetation} \\
     \midrule 
    MonoScene \cite{monoscene} & R101-DCN & C & 23.96 & 7.31 & 4.03 & 0.35 & 8.00 & 8.04 & 2.90 & 0.28 & 1.16 & 0.67 & 4.01 & 4.35 & 27.72 & 5.20 & 15.13 & 11.29 & 9.03 & 14.86\\
    Atlas \cite{atlas} & R101-DCN & C & 28.66 & 15.00 & 10.64 & 5.68 & 19.66 & 24.94 & 8.90 & 8.84 & 6.47 & 3.28 & 10.42 & 16.21 & 34.86 & 15.46 & 21.89 & 20.95 & 11.21 & 20.54 \\
    BEVFormer \cite{bevformer} & R101-DCN & C & 30.50 & 16.75 & 14.22 & 6.58 & 23.46 & 28.28 & 8.66 & 10.77 & 6.64 & 4.05 & 11.20 & 17.78 & 37.28 & 18.00 & 22.88 & 22.17 & 13.80 & 22.21 \\
    TPVFormer \cite{tpvformer} & R101-DCN & C & 30.86 & 17.10 & 15.96 & 5.31 & 23.86 & 27.32 & 9.79 & 8.74 & 7.09 & 5.20 & 10.97 & 19.22 & 38.87 & 21.25 & 24.26 & 23.15 & 11.73 & 20.81 \\
    C-CONet \cite{openoccupancy} & R101 & C & 26.10 & 18.40 & 18.60 & 10.00 & 26.40 & 27.40 & 8.60 & 15.70 & 13.30 & 9.70 & 10.90 & 20.20 & 33.00 & 20.70 & 21.40 & 21.80 & 14.70 & 21.30 \\
    InverseMatrixVT3D \cite{inversematrixvt3d} & R101-DCN & C & 31.85 & 18.88 & 18.39 & 12.46 & 26.30 & 29.11 & 11.00 & 15.74 & 14.78 & 11.38 & 13.31 & 21.61 & 36.30 & 19.97 & 21.26 & 20.43 & 11.49 & 18.47 \\
    RenderOcc \cite{renderocc} & R101 & C & 29.20 & 19.00 & 19.70 & 11.20 & 28.10 & 28.20 & 9.80 & 14.70 & 11.80 & 11.90 & 13.10 & 20.10 & 33.20 & 21.30 & 22.60 & 22.30 & 15.30 & 20.90 \\
    SurroundOcc \cite{surroundocc} & R101-DCN & C & 31.49 & 20.30 & 20.59 & 11.68 & 28.06 & 30.86 & 10.70 & 15.14 & 14.09 & 12.06 & 14.38 & 22.26 & 37.29 & \textbf{23.70} & 24.49 & 22.77 & 14.89 & 21.86 \\
    LMSCNet \cite{roldao2020lmscnet} & - & L & 36.60 & 14.90 & 13.10 & 4.50 & 14.70 & 22.10 & 12.60 & 4.20 & 7.20 & 7.10 & 12.20 & 11.50 & 26.30 & 14.30 & 21.10 & 15.20 & 18.50 & 34.20 \\
    L-CONet \cite{openoccupancy} & - & L & 39.40 & 17.70 & 19.20 & 4.00 & 15.10 & 26.90 & 6.20 & 3.80 & 6.80 & 6.00 & 14.10 & 13.10 & \textbf{39.70} & 19.10 & 24.00 & 23.90 & 25.10 & 35.70 \\
    M-CONet \cite{openoccupancy} & - & C+L & 39.20 & 24.70 & 24.80 & 13.00 & 31.60 & 34.80 & 14.60 & 18.00 & 20.00 & 14.70 & 20.00 & 26.60 & 39.20 & 22.80 & \textbf{26.10} & 26.00 & 26.00 & 37.10 \\
    \hline
    OccFusion & R101-DCN & C & 31.85 & 18.88 & 18.39 & 12.46 & 26.30 & 29.11 & 11.00 & 15.74 & 14.78 & 11.38 & 13.31 & 21.61 & 36.30 & 19.97 & 21.26 & 20.43 & 11.49 & 18.47 \\
    OccFusion & R101+VoxelNet & C+R & 33.97 & 20.73 & 20.46 & 13.98 & 27.99 & 31.52 & 13.68 & 18.45 & 15.79 & 13.05 & 13.94 & 23.84 & 37.85 & 19.60 & 22.41 & 21.20 & 16.16 & 21.81 \\
    OccFusion & R101+VoxelNet & C+L & 44.35 & 26.87 & 26.67 & 18.38 & 32.97 & 35.81 & 19.39 & 22.17 & 24.48 & \textbf{17.77} & 21.46 & 29.67 & 39.01 & 21.94 & 24.90 & \textbf{26.76} & 28.53 & 40.03 \\
    OccFusion & R101+VoxelNet & C+L+R & \textbf{44.66} & \textbf{27.30} & \textbf{27.09} & \textbf{19.56} & \textbf{33.68} & \textbf{36.23} & \textbf{21.66} & \textbf{24.84} & \textbf{25.29} & 16.33 & \textbf{21.81} & \textbf{30.01} & 39.53 & 19.94 & 24.94 & 26.45 & \textbf{28.93} & \textbf{40.41} \\
    \bottomrule
  \end{tabular}
  \end{adjustbox}
  \caption{\textbf{3D semantic occupancy prediction results on nuScenes validation set}. All methods are trained with dense occupancy labels from \cite{surroundocc}. Notion of modality: Camera (C), Lidar (L), Radar (R).}
  \label{occ}
\end{table*}

\begin{table*}[htpb]
  \centering
  \begin{adjustbox}{width=\textwidth}
  \begin{tabular}{c|c|c|c|ccccccccccccccccc}
    \toprule
    Method & Backbone & Input Modality & mIoU & 
    \rotatebox{90}{\textcolor{others}{$\bullet$} others} &
    \rotatebox{90}{\textcolor{barrier}{$\bullet$} barrier} & \rotatebox{90}{\textcolor{bicycle}{$\bullet$} bicycle} & \rotatebox{90}{\textcolor{bus}{$\bullet$} bus} & \rotatebox{90}{\textcolor{car}{$\bullet$} car} & \rotatebox{90}{\textcolor{construction}{$\bullet$} const. veh.} & \rotatebox{90}{\textcolor{motorcycle}{$\bullet$} motorcycle} & \rotatebox{90}{\textcolor{pedestrian}{$\bullet$} pedestrian} & \rotatebox{90}{\textcolor{cone}{$\bullet$} traffic cone} & \rotatebox{90}{\textcolor{trailer}{$\bullet$} trailer} & \rotatebox{90}{\textcolor{truck}{$\bullet$} truck} & \rotatebox{90}{\textcolor{driveable}{$\bullet$} drive. surf.} & \rotatebox{90}{\textcolor{flat}{$\bullet$} other flat} & \rotatebox{90}{\textcolor{sidewalk}{$\bullet$} sidewalk} & \rotatebox{90}{\textcolor{terrain}{$\bullet$} terrain} & \rotatebox{90}{\textcolor{manmade}{$\bullet$} manmade} & \rotatebox{90}{\textcolor{vegetation}{$\bullet$} vegetation} \\
     \midrule 
    MonoScene \cite{monoscene} & EfficientNetB7 & C & 6.06 & 1.75 & 7.23 & 4.26 & 4.93 & 9.38 & 5.67 & 3.98 & 3.01 & 5.90 & 4.45 & 7.17 & 14.91 & 6.32 & 7.92 & 7.43 & 1.01 & 7.65 \\
    BEVDet \cite{bevdet} & ResNet101 & C & 11.73 & 2.09 & 15.29 & 0.0 & 4.18 & 12.97 & 1.35 & 0.0 & 0.43 & 0.13 & 6.59 & 6.66 & 52.72 & 19.04 & 26.45 & 21.78 & 14.51 & 15.26 \\
    BEVFormer \cite{bevformer} & ResNet101 & C & 23.67 & 5.03 & 38.79 & 9.98 & 34.41 & 41.09 & 13.24 & 16.50 & 18.15 & 17.83 & 18.66 & 27.70 & 48.95 & 27.73 & 29.08 & 25.38 & 15.41 & 14.46 \\
    BEVStereo \cite{bevstereo} & ResNet101 & C & 24.51 & 5.73 & 38.41 & 7.88 & 38.70 & 41.20 & 17.56 & 17.33 & 14.69 & 10.31 & 16.84 & 29.62 & 54.08 & 28.92 & 32.68 & 26.54 & 18.74 & 17.49 \\
    TPVFormer \cite{tpvformer} & ResNet101 & C & 28.34 & 6.67 & 39.20 & 14.24 & 41.54 & 46.98 & 19.21 & 22.64 & 17.87 & 14.54 & 30.20 & 35.51 & 56.18 & 33.65 & 35.69 & 31.61 & 19.97 & 16.12 \\
    OccFormer \cite{occformer} & ResNet101 & C & 21.93 & 5.94 & 30.29 & 12.32 & 34.40 & 39.17 & 14.44 & 16.45 & 17.22 & 9.27 & 13.90 & 26.36 & 50.99 & 30.96 & 34.66 & 22.73 & 6.76 & 6.97 \\
    CTF-Occ \cite{occ3d} & ResNet101 & C & 28.53 & 8.09 & 39.33 & 20.56 & 38.29 & 42.24 & 16.93 & 24.52 & 22.72 & 21.05 & 22.98 & 31.11 & 53.33 & 33.84 & 37.98 & 33.23 & 20.79 & 18.00 \\
    RenderOcc \cite{renderocc} & ResNet101 & C & 26.11 & 4.84 & 31.72 & 10.72 & 27.67 & 26.45 & 13.87 & 18.20 & 17.67 & 17.84 & 21.19 & 23.25 & 63.20 & 36.42 & 46.21 & 44.26 & 19.58 & 20.72 \\
    BEVDet4D \cite{bevdet4d}* & Swin-B & C & 42.02 & 12.15 & 49.63 & 25.10 & 52.02 & 54.46 & 27.87 & 27.99 & 28.94 & 27.23 & 36.43 & 42.22 & 82.31 & 43.29 & 54.46 & 57.90 & 48.61 & 43.55 \\
    PanoOcc \cite{panoocc}* & ResNet101 & C & 42.13 & 11.67 & 50.48 & 29.64 & 49.44 & 55.52 & 23.29 & 33.26 & 30.55 & 30.99 & 34.43 & 42.57 & 83.31 & 44.23 & 54.40 & 56.04 & 45.94 & 40.40 \\
    FB-OCC \cite{fbocc}* & ResNet101 & C & 43.41 & 12.10 & 50.23 & \textbf{32.31} & 48.55 & 52.89 & 31.20 & 31.25 & 30.78 & 32.33 & 37.06 & 40.22 & \textbf{83.34} & \textbf{49.27} & \textbf{57.13} & \textbf{59.88} & 47.67 & 41.76 \\
    OctreeOcc \cite{octreeocc}* & ResNet101 & C & 44.02 & 11.96 & \textbf{51.70} & 29.93 & 53.52 & 56.77 & 30.83 & 33.17 & 30.65 & 29.99 & 37.76 & 43.87 & 83.17 & 44.52 & 55.45 & 58.86 & 49.52 & 46.33 \\
    \hline
    OccFusion* & R101+VoxelNet & C+R  & 38.26 & 10.11 & 43.48 & 26.63 & 48.25 & 53.29 & 24.98 & 33.35 & 31.75 & 27.31 & 29.88 & 40.15 & 77.59 & 35.47 & 43.64 & 48.01 & 42.73 & 33.81 \\
    OccFusion* & R101+VoxelNet & C+L  & \textbf{46.79} & 11.65 & 47.81 & 32.07 & 57.27 & 57.51 & 31.80 & 40.11 & \textbf{47.35} & \textbf{33.74} & \textbf{45.81} & \textbf{50.35} & 78.79 & 37.17 & 44.36 & 53.36 & 63.18 & 63.20 \\
    OccFusion* & R101+VoxelNet & C+L+R & 46.67 & \textbf{12.37} & 50.33 & 31.53 & \textbf{57.62} & \textbf{58.81} & \textbf{33.97} & \textbf{41.00} & 47.18 & 29.67 & 42.03 & 48.04 & 78.39 & 35.68 & 47.26 & 52.74 & \textbf{63.46} & \textbf{63.30} \\
    \bottomrule
  \end{tabular}
  \end{adjustbox}
  \caption{\textbf{3D semantic occupancy prediction results on Occ3D benchmark}. All methods are trained with dense occupancy labels from \cite{occ3d}. Notion of modality: Camera (C), Lidar (L), Radar (R). "*" denotes training with the camera mask.}
  \label{occ_3d}
\end{table*}

\begin{table*}[t]
  \centering
  \begin{adjustbox}{width=\textwidth}
  \begin{tabular}{c|cc|ccccccccccccccccccc}
    \toprule
    Method & IoU & mIoU & \rotatebox{90}{road} & \rotatebox{90}{sidewalk} & \rotatebox{90}{parking} & \rotatebox{90}{other-ground} & \rotatebox{90}{building} & \rotatebox{90}{car} & \rotatebox{90}{truck} & \rotatebox{90}{bicycle} & \rotatebox{90}{motorcycle} & \rotatebox{90}{other-vehicle} & \rotatebox{90}{vegetation} & \rotatebox{90}{trunk} & \rotatebox{90}{terrain} & \rotatebox{90}{person} & \rotatebox{90}{bicyclist} & \rotatebox{90}{motorcyclist} & \rotatebox{90}{fence} & \rotatebox{90}{pole} & \rotatebox{90}{traf.sign} \\
     \midrule 
    LMSCNet \cite{roldao2020lmscnet} & 28.61 
    & 6.70 & 40.68 & 18.22 & 4.38 & 0.00 & 10.31 & 18.33 & 0.00 & 0.00 & 0.00 & 0.00 & 13.66 & 0.02 & 20.54 & 0.00 & 0.00 & 0.00 & 1.21 & 0.00 & 0.00 \\
    AICNet \cite{li2020anisotropic} & 29.59 
    & 8.31 & 43.55 & 20.55 & 11.97 & 0.07 & 12.94 & 14.71 & 4.53 & 0.00 & 0.00 & 0.00 & 15.37 & 2.90 & 28.71 & 0.00 & 0.00 & 0.00 & 2.52 & 0.06 & 0.00 \\
    3DSketch \cite{3dsketch} & 33.30 
    & 7.50 & 41.32 & 21.63 & 0.00 & 0.00 & 14.81 & 18.59 & 0.00 & 0.00 & 0.00 & 0.00 & 19.09 & 0.00 & 26.40 & 0.00 & 0.00 & 0.00 & 0.73 & 0.00 & 0.00 \\
    JS3C-Net \cite{JS3CNet} & 38.98 
    & 10.31 & 50.49 & 23.74 & 11.94 & 0.07 & 15.03 & 24.65 & 4.41 & 0.00 & 0.00 & 6.15 & 18.11 & 4.33 & 26.86 & 0.67 & 0.27 & 0.00 & 3.94 & 3.77 & 1.45 \\
    MonoScene \cite{monoscene} & 36.86 
    & 11.08 & 56.52 & 26.72 & 14.27 & 0.46 & 14.09 & 23.26 & 6.98 & 0.61 & 0.45 & 1.48 & 17.89 & 2.81 & 29.64 & 1.86 & 1.20 & 0.00 & 5.84 & 4.14 & 2.25 \\
    TPVFormer \cite{tpvformer} & 35.61 
    & 11.36 & 56.50 & 25.87 & 20.60 & 0.85 & 13.88 & 23.81 & 8.08 & 0.36 & 0.05 & 4.35 & 16.92 & 2.26 & 30.38 & 0.51 & 0.89 & 0.00 & 5.94 & 3.14 & 1.52\\
    InverseMatrixVT3D \cite{inversematrixvt3d} & 36.22 
    & 11.81 & 52.99 & 25.84 & 20.04 & 0.09 & 13.17 & 24.08 & 10.25 & 1.85 & 2.65 & 6.80 & 16.98 & 3.09 & 27.77 & \textbf{4.01} & 3.13 & 0.00 & 4.94 & 4.05 & 2.67 \\
    VoxFormer \cite{li2023voxformer} & 44.02 
    & 12.35 & 54.76 & 26.35 & 15.50 & 0.70 & 17.65 & 25.79 & 5.63 & 0.59 & 0.51 & 3.77 & 24.39 & 5.08 & 29.96 & 1.78 & 3.32 & 0.00 & 7.64 & 7.11 & 4.18\\
    OccFormer \cite{occformer} & 36.50 
    & 13.46 & 58.84 & 26.88 & 19.61 & 0.31 & 14.40 & 25.09 & 25.53 & 0.81 & 1.19 & 8.52 & 19.63 & 3.93 & 32.63 & 2.78 & 2.82 & 0.00 & 5.61 & 4.26 & 2.86\\
    Symphonies \cite{jiang2023symphonize} & 41.44 
    & 13.44 & 55.78 & 26.77 & 14.57 & 0.19 & 18.76 & 27.23 & 15.99 & 1.44 & 2.28 & 9.52 & 24.50 & 4.32 & 28.49 & 3.19 & \textbf{8.09} & 0.00 & 6.18 & 8.99 & 5.39 \\
    OctreeOcc \cite{octreeocc} & 44.71 
    & 13.12 & 55.13 & 26.74 & 18.68 & 0.65 & 18.69 & 28.07 & 16.43 & 0.64 & 0.71 & 6.03 & 25.26 & 4.89 & 32.47 & 2.25 & 2.57 & 0.00 & 4.01 & 3.72 & 2.36 \\
    UDNet \cite{udnet} & \textbf{58.90} & 20.70 & \textbf{67.00} & \textbf{37.20} & 20.30 & \textbf{2.20} & 36.00 & 42.10 & \textbf{25.70} & 1.80 & 2.30 & \textbf{11.20} & 40.10 & 18.30 & \textbf{45.80} & 2.50 & 1.20 & 0.00 & 11.90 & 23.00 & 3.80 \\
    \hline
    OccFusion(C+L) & 58.68
    & \textbf{21.92} & 65.67 & 36.33 & \textbf{23.08} & 0.00 & \textbf{39.09} & \textbf{45.62} & 20.05 & \textbf{2.96} & \textbf{3.51} & 8.76 & \textbf{40.68} & \textbf{19.37} & 45.53 & 3.16 & 4.37 & 0.00 & \textbf{15.70} & \textbf{27.57} & \textbf{15.21} \\
    \bottomrule
  \end{tabular}
  \end{adjustbox}
  \caption{\textbf{3D semantic scene completion performance on SemanticKITTI validation set}. Notion of modality: Camera (C), Lidar (L)}
  \label{kitti}
\end{table*}

\subsection{Challenging Scenarios Performance Analysis}
We assess various sensor fusion strategies in challenging nighttime and rainy scenarios to gain deeper insights into sensor fusion properties and effectiveness. The performance of the models in these scenarios is presented in Table \ref{occ_rainy} and Table \ref{occ_night}. 

In the rainy scenario, despite the sparse 3D point cloud provided by the radar sensors, we observed 2\% performance gain by integrating surround-view cameras with radar. Furthermore, despite its reflection issue in rainy scenarios, the lidar sensor can still substantially contribute to the model's overall performance owing to its dense 3D point cloud. Another reason is the absence of severe rainy conditions in the nuScenes dataset. The dataset primarily includes light to moderate rain scenarios, wherein the lidar data maintains a consistently high quality. Our model performs best by integrating the information from all three sensors.

In nighttime scenarios, it is undeniable that purely vision-centric approaches perform poorly due to the sensitivity of surround-view cameras to varying illumination conditions. We found that integrating information from surround-view radars notably augmented the model's performance, leading to an approximate 1.2\% enhancement. Furthermore, including radar data significantly improves the capability to predict dynamic objects. Particularly, a performance boost of around 4\% is observed for the car class. Likewise, we note performance gains of approximately 0.7\% for bicycles and 3\% for motorcycles as small dynamic objects. This progress is linked to the velocity measurement function of surround-view radars and demonstrates the effective integration of these features with camera attributes within our framework.
What's more, including lidar information yielded an additional performance gain of 4.7\% 

\begin{table*}[htpb]
  \centering
  \begin{adjustbox}{width=\textwidth}
  \begin{tabular}{c|c|c|cc|cccccccccccccccc}
    \toprule
    Method & Backbone & Input Modality & IoU & mIoU & \rotatebox{90}{\textcolor{barrier}{$\bullet$} barrier} & \rotatebox{90}{\textcolor{bicycle}{$\bullet$} bicycle} & \rotatebox{90}{\textcolor{bus}{$\bullet$} bus} & \rotatebox{90}{\textcolor{car}{$\bullet$} car} & \rotatebox{90}{\textcolor{construction}{$\bullet$} const. veh.} & \rotatebox{90}{\textcolor{motorcycle}{$\bullet$} motorcycle} & \rotatebox{90}{\textcolor{pedestrian}{$\bullet$} pedestrian} & \rotatebox{90}{\textcolor{cone}{$\bullet$} traffic cone} & \rotatebox{90}{\textcolor{trailer}{$\bullet$} trailer} & \rotatebox{90}{\textcolor{truck}{$\bullet$} truck} & \rotatebox{90}{\textcolor{driveable}{$\bullet$} drive. surf.} & \rotatebox{90}{\textcolor{flat}{$\bullet$} other flat} & \rotatebox{90}{\textcolor{sidewalk}{$\bullet$} sidewalk} & \rotatebox{90}{\textcolor{terrain}{$\bullet$} terrain} & \rotatebox{90}{\textcolor{manmade}{$\bullet$} manmade} & \rotatebox{90}{\textcolor{vegetation}{$\bullet$} vegetation} \\
     \midrule 
    OccFusion & R101-DCN & C & 31.10 & 18.99 & 18.55 & 14.29 & 22.28 & 30.02 & 10.19 & 15.20 & 10.03 & 9.71 & 13.28 & 20.98 & 37.18 & 23.47 & 27.74 & 17.46 & 10.36 & 23.13 \\
    SurroundOcc \cite{surroundocc} & R101-DCN & C & 30.57 & 19.85 & 21.40 & 12.75 & 25.49 & 31.31 & 11.39 & 12.65 & 8.94 & 9.48 & 14.51 & 21.52 & 35.34 & \textbf{25.32} & 29.89 & 18.37 & 14.44 & 24.78 \\
    OccFusion & R101+VoxelNet & C+R & 33.75 & 20.78 & 20.14 & 16.33 & 26.37 & 32.39 & 11.56 & 17.08 & 11.14 & 10.54 & 13.61 & 22.42 & 37.50 & 22.79 & 29.50 & 17.58 & 17.06 & 26.49 \\
    OccFusion & R101+VoxelNet & C+L & 43.36 & 26.55 & 24.95 & \textbf{19.11} & \textbf{34.23} & 36.07 & 17.01 & 21.07 & 18.87 & \textbf{17.46} & 21.81 & 28.73 & 37.82 & 24.39 & \textbf{30.80} & \textbf{20.37} & 28.95 & 43.12 \\
    OccFusion & R101+VoxelNet & C+L+R & \textbf{43.50} & \textbf{26.72} & \textbf{25.30} & 18.71 & 33.58 & \textbf{36.28} & \textbf{17.76} & \textbf{22.44} & \textbf{20.80} & 15.89 & \textbf{22.63} & \textbf{28.75} & \textbf{39.28} & 22.72 & 30.78 & 20.15 & \textbf{28.99} & \textbf{43.37} \\
    \bottomrule
  \end{tabular}
  \end{adjustbox}
  \caption{\textbf{3D semantic occupancy prediction results on nuScenes validation rainy scenario subset}. All methods are trained with dense occupancy labels from \cite{surroundocc}. Notion of modality: Camera (C), Lidar (L), Radar (R).}
  \label{occ_rainy}
\end{table*}

\begin{table*}[htpb]
  \centering
  \begin{adjustbox}{width=\textwidth}
  \begin{tabular}{c|c|c|cc|cccccccccccccccc}
    \toprule
    Method & Backbone & Input Modality & IoU & mIoU & \rotatebox{90}{\textcolor{barrier}{$\bullet$} barrier} & \rotatebox{90}{\textcolor{bicycle}{$\bullet$} bicycle} & \rotatebox{90}{\textcolor{bus}{$\bullet$} bus} & \rotatebox{90}{\textcolor{car}{$\bullet$} car} & \rotatebox{90}{\textcolor{construction}{$\bullet$} const. veh.} & \rotatebox{90}{\textcolor{motorcycle}{$\bullet$} motorcycle} & \rotatebox{90}{\textcolor{pedestrian}{$\bullet$} pedestrian} & \rotatebox{90}{\textcolor{cone}{$\bullet$} traffic cone} & \rotatebox{90}{\textcolor{trailer}{$\bullet$} trailer} & \rotatebox{90}{\textcolor{truck}{$\bullet$} truck} & \rotatebox{90}{\textcolor{driveable}{$\bullet$} drive. surf.} & \rotatebox{90}{\textcolor{flat}{$\bullet$} other flat} & \rotatebox{90}{\textcolor{sidewalk}{$\bullet$} sidewalk} & \rotatebox{90}{\textcolor{terrain}{$\bullet$} terrain} & \rotatebox{90}{\textcolor{manmade}{$\bullet$} manmade} & \rotatebox{90}{\textcolor{vegetation}{$\bullet$} vegetation} \\
     \midrule 
    OccFusion & R101-DCN & C & 24.49 & 9.99 & 10.40 & 12.03 & 0.00 & 29.94 & 0.00 & 9.92 & 4.88 & 0.91 & 0.00 & 17.79 & 29.10 & 2.37 & 10.80 & 9.40 & 8.68 & 13.57 \\
    SurroundOcc \cite{surroundocc} & R101-DCN & C & 24.38 & 10.80 & 10.55 & \textbf{14.60} & 0.00 & 31.05 & 0.00 & 8.26 & 5.37 & 0.58 & 0.00 & 18.75 & 30.72 & \textbf{2.74} & 12.39 & 11.53 & 10.52 & 15.77 \\
    OccFusion & R101+VoxelNet & C+R & 27.09 & 11.13 & 10.78 & 12.77 & 0.00 & 33.50 & 0.00 & 12.72 & 4.91 & 0.61 & 0.00 & 19.97 & 29.51 & 0.94 & 12.15 & 10.72 & 11.81 & 17.72 \\
    OccFusion & R101+VoxelNet & C+L & 41.38 & 15.26 & 12.74 & 13.52 & 0.00 & 35.85 & 0.00 & 15.33 & \textbf{13.19} & 0.83 & 0.00 & 23.78 & \textbf{32.49} & 0.92 & 14.24 & 20.54 & 23.57 & \textbf{37.10} \\
    OccFusion & R101+VoxelNet & C+L+R & \textbf{41.47} & \textbf{15.82} & \textbf{13.27} & 13.53 & 0.00 & \textbf{36.41} & 0.00 & \textbf{19.71} & 12.16 & \textbf{2.04} & 0.00 & \textbf{25.90} & 32.44 & 0.80 & \textbf{14.30} & \textbf{21.06} & \textbf{24.49} & 37.00 \\
    \bottomrule
  \end{tabular}
  \end{adjustbox}
  \caption{\textbf{3D semantic occupancy prediction results on nuScenes validation night scenario subset}. All methods are trained with dense occupancy labels from \cite{surroundocc}. Notion of modality: Camera (C), Lidar (L), Radar (R).}
  \label{occ_night}
\end{table*}

\subsection{Perception Range Impact On Model Performance} 
Multi-sensor fusion improves the final model's robustness to illumination and weather conditions and extends the model's perception range. We take the vehicle's centre as the origin and $R$ as the radius. 
By adjusting the length of $R$, we study the characteristics of different sensor fusion strategies under different perception ranges in different scenarios. We evaluate each model and different sensor fusion strategies at $R=[20m,25m,30m,35m,40m,45m,50m]$.

\begin{figure*}[t]
     \centering
     \begin{subfigure}[]{0.32\textwidth}
         \centering
         \includegraphics[width=\textwidth]{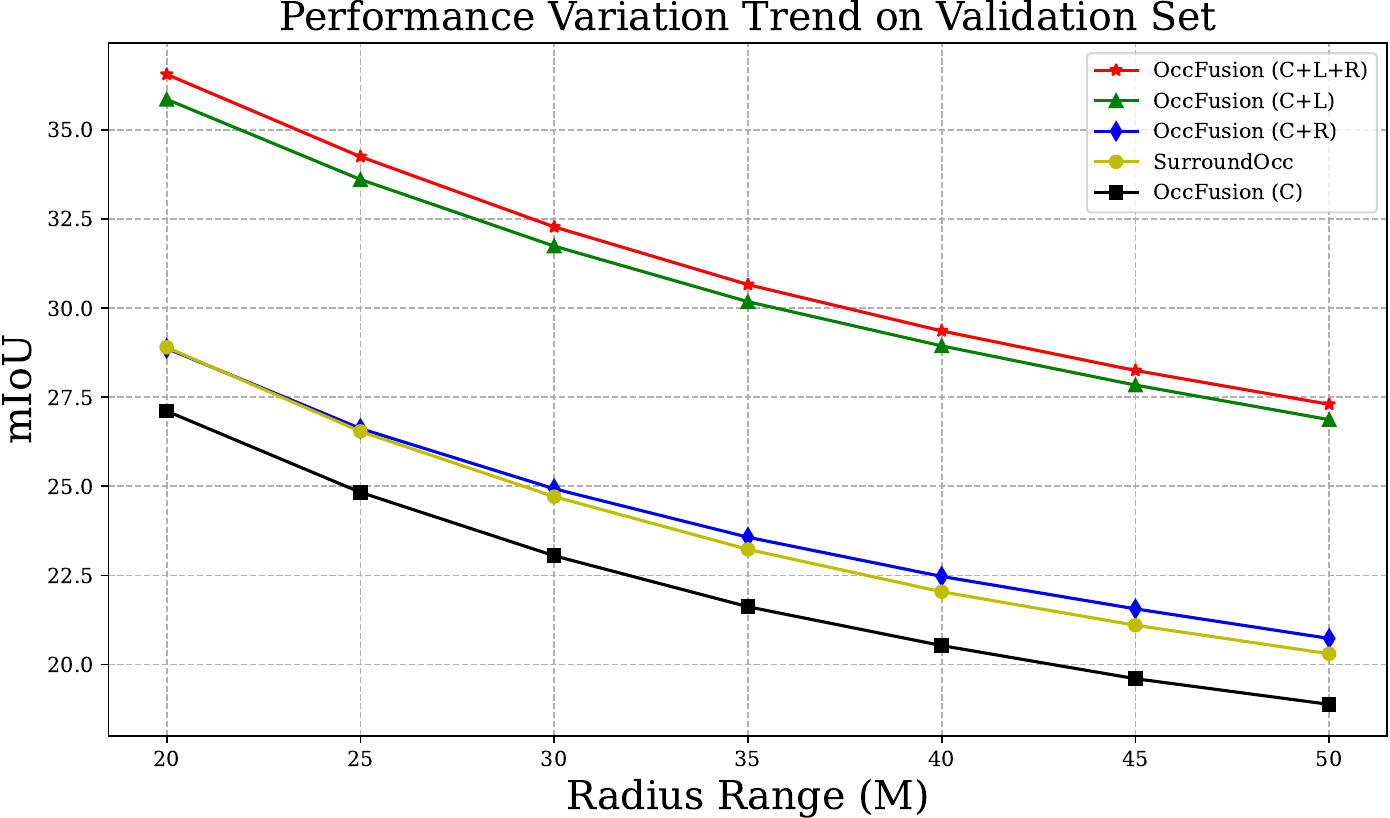}
         \caption{}
         \label{whole_trend}
     \end{subfigure}
     \begin{subfigure}[]{0.33\textwidth}
         \centering
         \includegraphics[width=\textwidth]{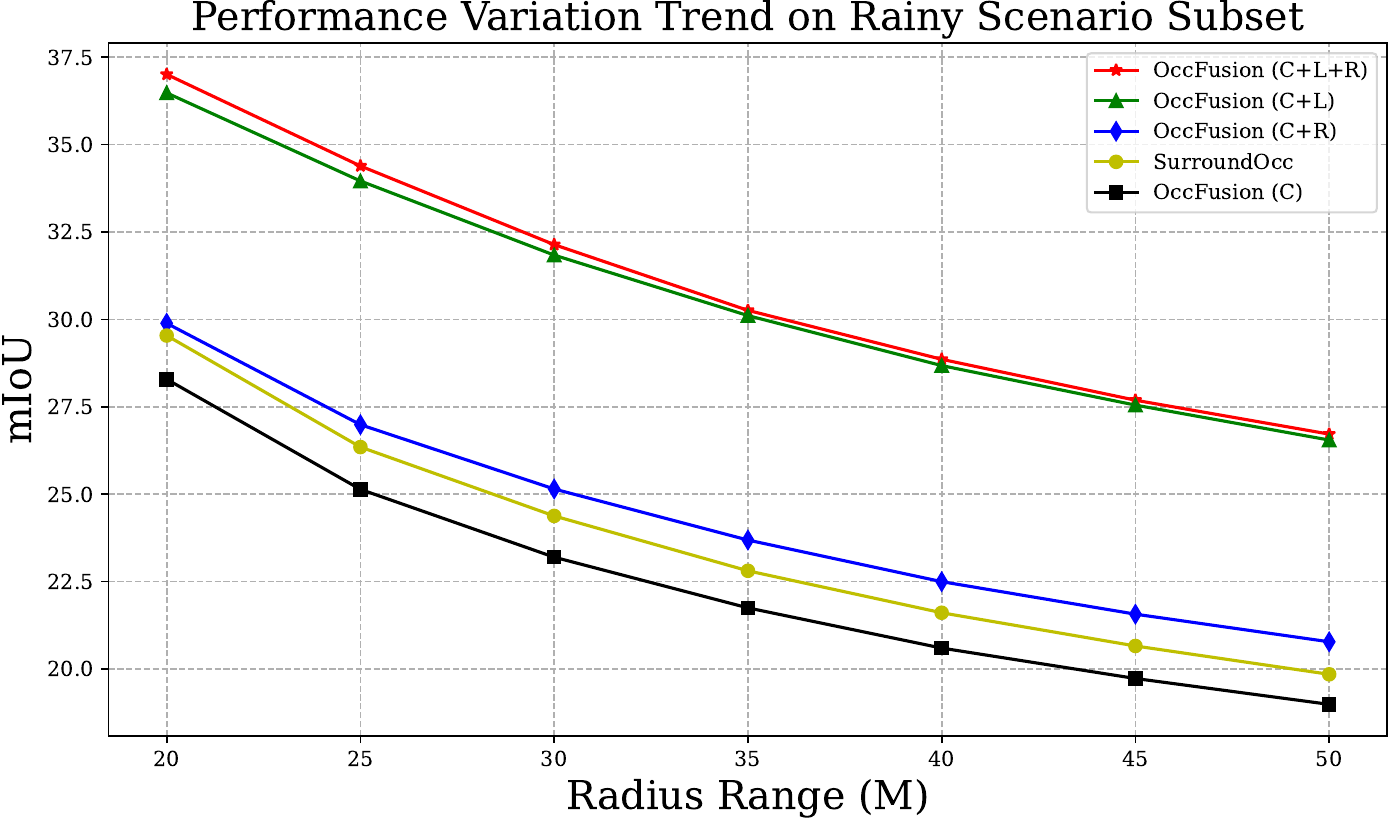}
         \caption{}
         \label{rainy_trend}
     \end{subfigure}
     \begin{subfigure}[]{0.33\textwidth}
         \centering
         \includegraphics[width=\textwidth]{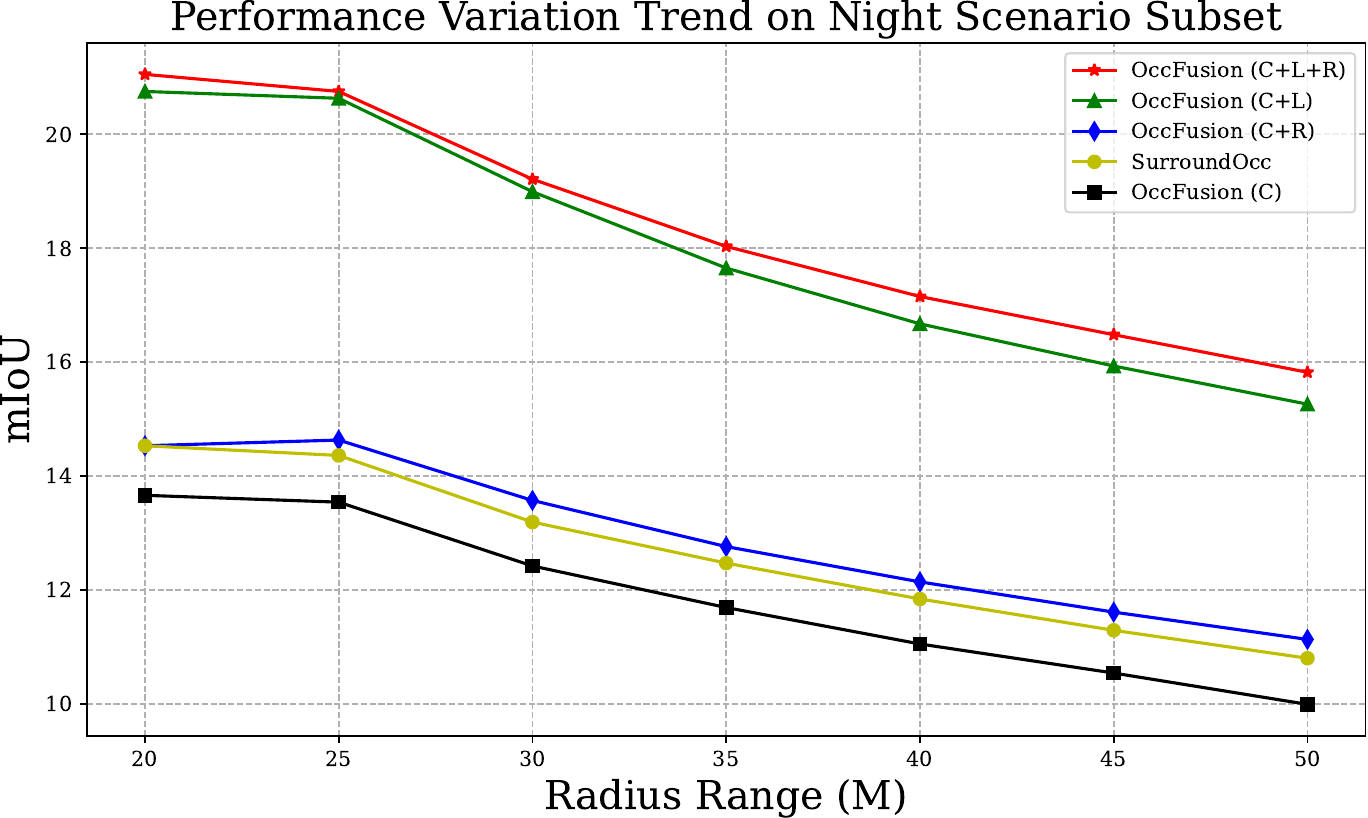}
         \caption{}
         \label{night_trend}
     \end{subfigure}
        \caption{\small Performance variation trend for 3D semantic occupancy prediction task. (a) mIoU performance variation trend on the whole nuScenes validation set, (b) mIoU performance variation trend on the nuScenes validation rainy scenario subset, and (c) mIoU performance variation on the nuScenes validation night scenario subset.\textbf{Better viewed when zoomed in.}} 
        \label{perform_var}
\end{figure*}

The performance variation trend of each model concerning $R$ on the nuScenes validation set is depicted in Figure \ref{whole_trend}. Our model achieves significantly improved performance by integrating radar and lidar data, particularly at longer ranges.

Figure \ref{rainy_trend} depicts the performance variation trend in the rainy scenario. The performance trend of OccFusion(C) displays a notable discrepancy compared to OccFusion(C+R), and this difference becomes more pronounced as the perception range expands. This phenomenon demonstrates the radar sensor's enhancement to the purely vision-centric algorithms.  Nevertheless, the variation trend of OccFusion(C+L) only slightly deviates from that of OccFusion(C+L+R) when the radius exceeds 30m, which shows that in the presence of integrated lidar data in rainy scenes, as the sensing range expands, the contribution of radar decreases.

In the nighttime scenario, the variation trend of performance is shown in Figure \ref{night_trend}. It demonstrates that adding radar information significantly enhances the framework's ability to perceive objects at extended distances. Interestingly, the performance disparity between OccFusion (C+L+R) and OccFusion (C+L) widens as the perception range increases, opposite to the trend observed in the rainy scenario.

\subsection{Framework Qualitative Analysis}
\begin{figure*}[htbp]
\centering
\begin{subfigure}[]{\textwidth}
\includegraphics[width=\textwidth]{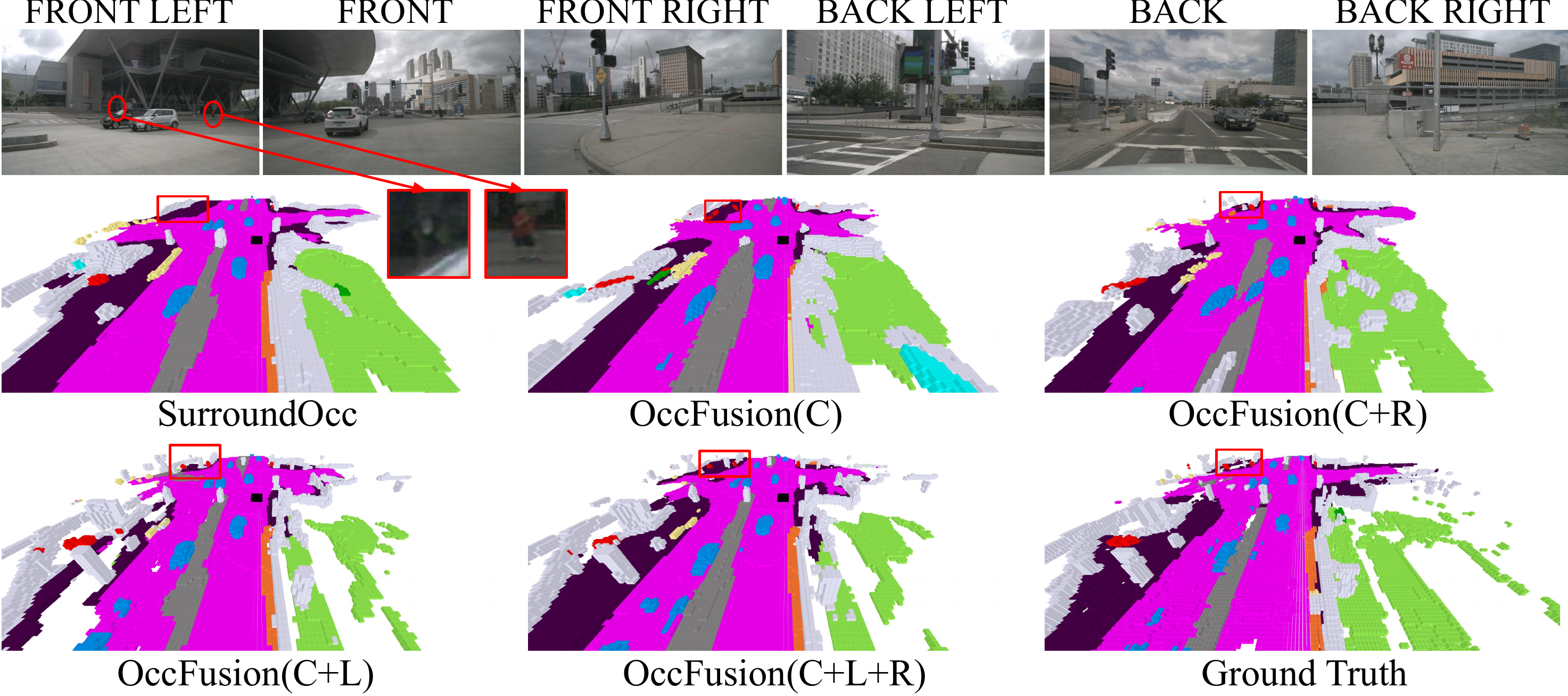}\\
\includegraphics[width=\textwidth]{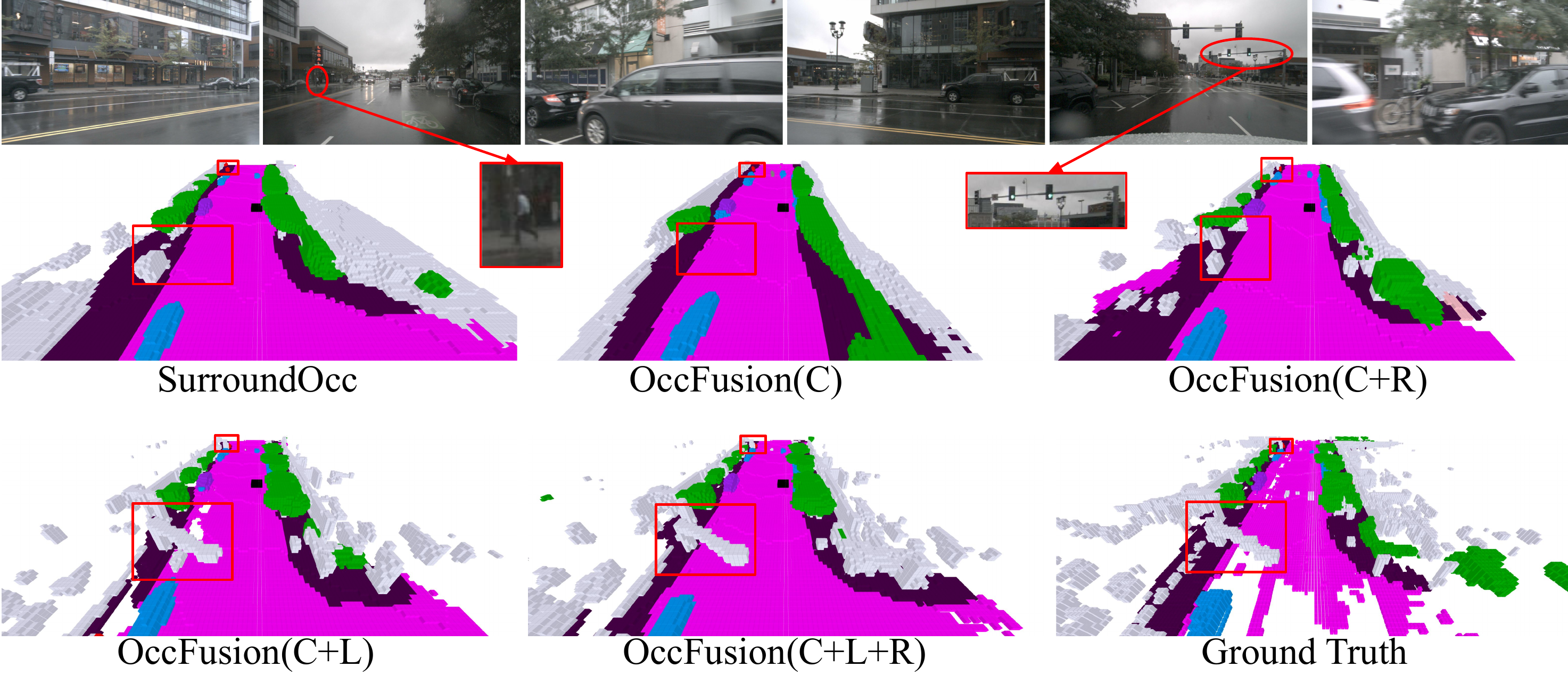}\\
\includegraphics[width=\textwidth]{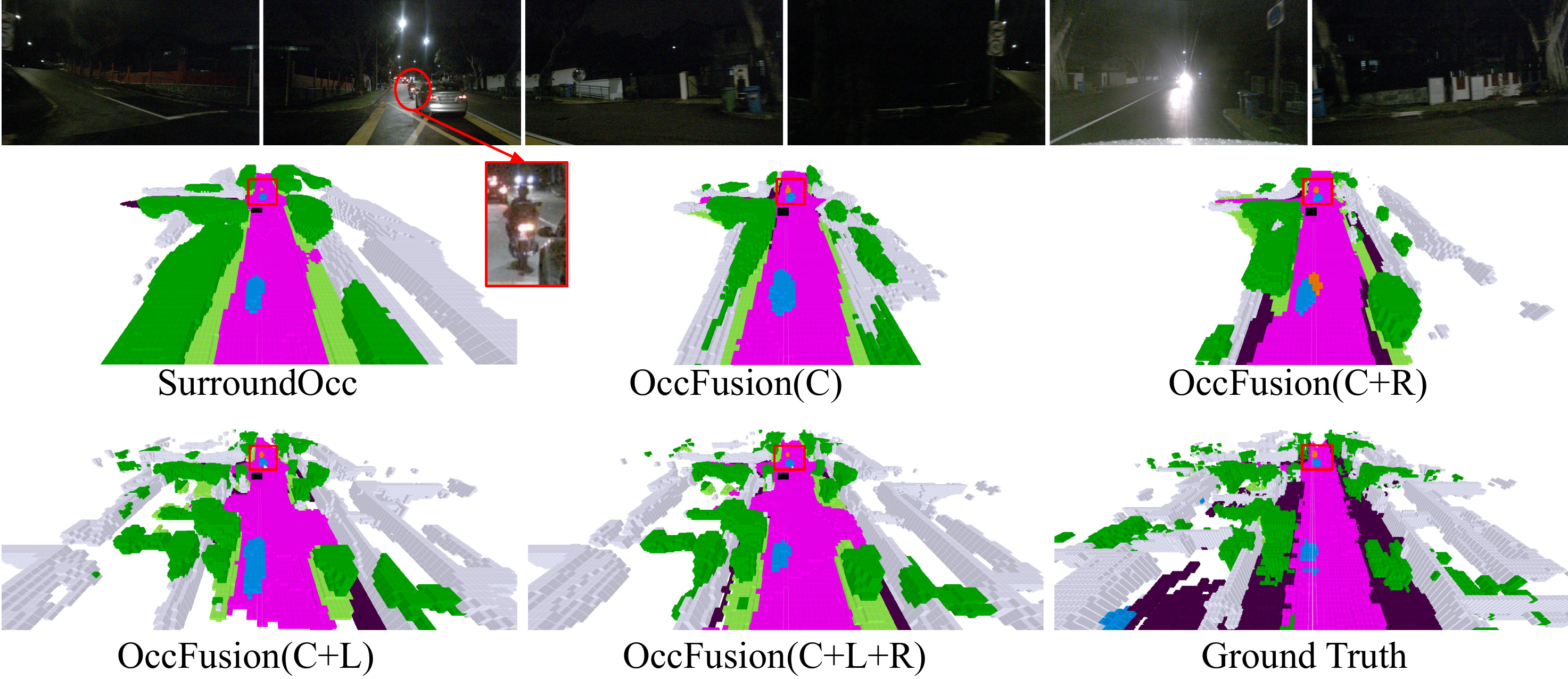}
\end{subfigure}
    

\caption{Qualitative results for daytime, rainy, and nighttime scenarios displayed in the upper, middle, and bottom sections, respectively. \textbf{Better viewed when zoomed in.} Notion of modality: Camera (C), Lidar (L), Radar (R).}
\label{qualitative}
\end{figure*}
We conducted qualitative analysis by generating visualizations of recent SOTA algorithms and comparing them with the prediction results from our framework with different sensor fusion strategies. The overall visualization result is demonstrated in Figure \ref{qualitative}. In Figure \ref{qualitative} upper, we present the prediction results for the daytime scenario; in the middle, we show the prediction results for the rainy scenario; and at the bottom, we present the prediction results for the nighttime scenario. The red rectangle highlights the main discrepancy of each prediction result under each scenario.

In the daytime scenario, as shown in Figure \ref{qualitative} upper, algorithms that rely solely on surround-view cameras cannot accurately predict pedestrians at remote distances, either failing to identify them or misestimating their numbers. This issue has been partially resolved by integrating radar information with the camera, which means the radar data helps the model extend its perception range. Furthermore, lidar information has further enhanced the framework's ability to model the 3D world, particularly in capturing the geometry and contour of static objects.

In the rainy scenario, as shown in Figure \ref{qualitative} middle, vision-centric algorithms have trouble predicting remote range overlay objects. This issue has been mitigated by merging radar information with cameras, but models still have trouble predicting the manmade buildings that hang in the sky. This issue can be solved by adding lidar data to the model. This visualization result reveals that three-sensor fusion extends the model perception range and enhances the model's 3D world structure detail-capturing capability.

In the nighttime scenario, the surround-view cameras are susceptible to illumination changes and perform poorly in dim environments. As a result, it is not surprising that purely vision-centric algorithms yield terrible prediction results in such scenarios, as shown in Figure \ref{qualitative} bottom. Surprisingly, we found that even by adding only radar data, which provides a sparse 3D point cloud, the model significantly improves predicting static objects such as vegetation and manmade structures. Furthermore, when lidar data is merged, the model's prediction results improve significantly. It is worth noting that even after merging lidar data, the OccFusion (C+L+R) model fails to classify the nearby sidewalk. This phenomenon can be attributed to the fact that lidar sensors do not provide rich semantic information, and in this particular scenario, the camera's semantic information is also significantly degraded.

\subsection{Framework Training Convergence Speed Study}
\begin{figure}[htbp]
\centering
{\includegraphics[width=0.95\columnwidth]{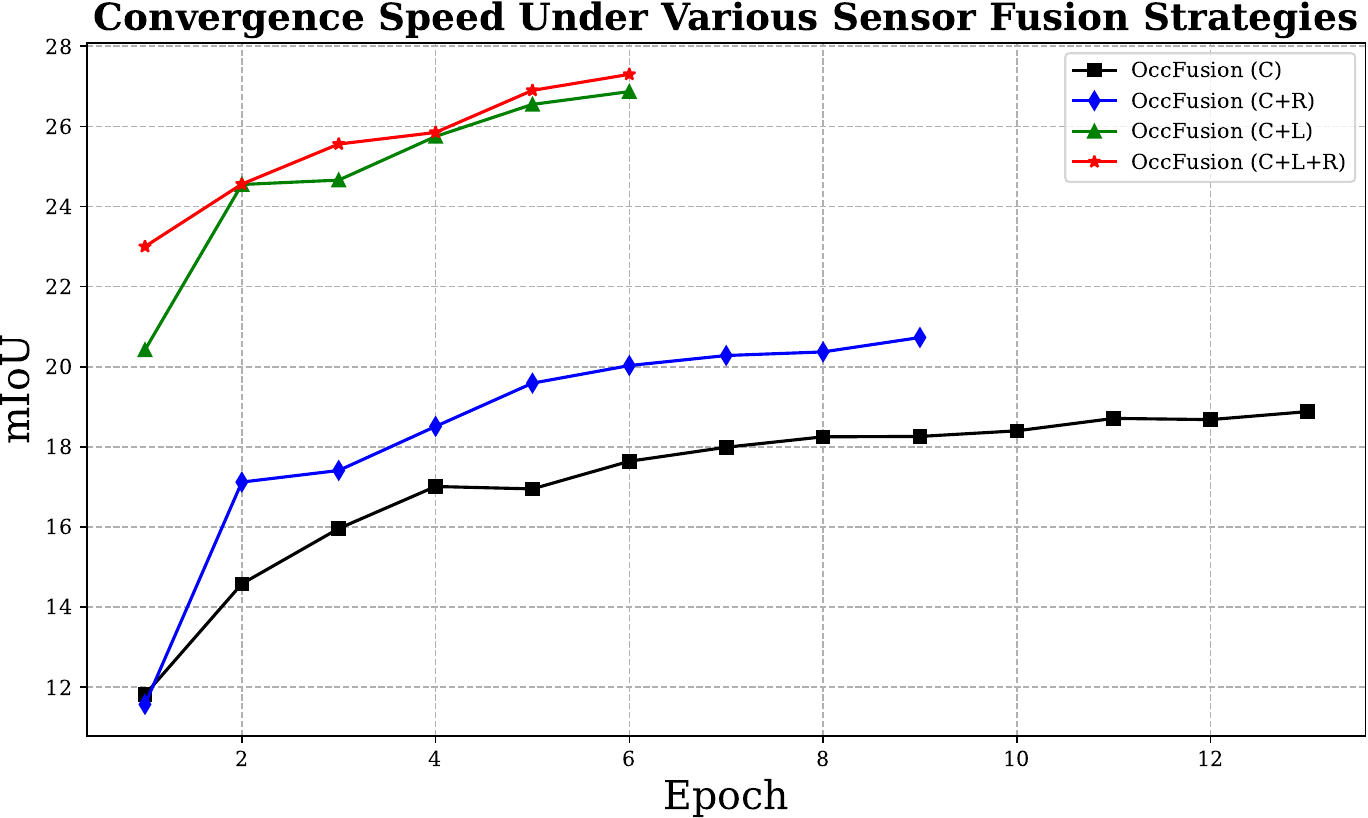}}
\caption{The convergence speed of the framework during the training phase under various sensor fusion strategies. Notion of modality: Camera (C), Lidar (L), Radar (R).}\label{conv_speed}
\end{figure}
During the training phase, we observed that different sensor fusion strategies influence the framework's ultimate performance and notably impact its overall training duration. As illustrated in Figure \ref{conv_speed}, OccFusion (C), a vision-centric method, required 13 training epochs to reach optimal performance. In contrast, OccFusion (C+R), integrating radar information with cameras, reduced the total training epochs to 9 and achieved approximately a 2\% performance enhancement. Combining Lidar information with cameras further reduced the training epochs to 6, resulting in a performance improvement of 6\%. This occurrence highlights the benefits of sensor fusion, which enhances the final framework performance and accelerates its convergence speed during training.

\subsection{Framework Efficiency Study}
We assess the effectiveness of each sensor fusion strategy implemented in our framework and compare them with other SOTA algorithms. Table \ref{efficiency} provides detailed information on the framework's efficiency. By incorporating more sensor information, our framework becomes more complex and necessitates more trainable parameters. Consequently, this leads to increased GPU memory utilisation and higher latency during inference.
\begin{table}[h]
  \centering
  \begin{adjustbox}{width=\columnwidth}
  \begin{tabular}{c|ccc}
    \toprule
    {Method} & {Latency (ms) ($\downarrow$)} & {Memory (GB) ($\downarrow$)} & Params \\
    \midrule
    SurroundOcc \cite{surroundocc} & 472 & 5.98 & 180.51M \\
    InverseMatrixVT3D \cite{inversematrixvt3d} & 447 & 4.41 & 67.18M \\
    OccFusion(C+R) & 588 & 5.56 & 92.71M \\
    OccFusion(C+L) & 591 & 5.56 & 92.71M \\
    OccFusion(C+L+R) & 601 & 5.78 & 114.97M \\
    \bottomrule
  \end{tabular}
  \end{adjustbox}
  \caption{Model efficiency comparison of different methods. The experiments are performed on a single A10 using six multi-camera images, lidar, and radar data. For input image resolution, all methods adopt $1600\times900$. $\downarrow$:the lower, the better.}
  \label{efficiency}
\end{table}

\subsection{Framework Ablation Study}
\subsubsection{Ablation on multi-scale mechanism}
We investigated the influence of multi-level supervision and a multi-scale coarse-to-fine feature refinement structure on the overall performance. The ablation study result is shown in Table \ref{multi}.
\begin{table}[h]
  \centering
  \begin{adjustbox}{width=0.7\columnwidth}
  \begin{tabular}{cc|c|cc}
    \toprule
    Multi.Stru & Multi.Sup & Params & mIoU$\uparrow$ & IoU$\uparrow$ \\
    \midrule
    \Checkmark & \Checkmark & 114.97M & 23.00\% & 43.90\% \\
    \Checkmark & \XSolidBrush & 114.97M & 17.79\% & 43.78\% \\
    \XSolidBrush & \XSolidBrush & 54.99M & 19.47\% & 42.46\% \\ 
    \bottomrule
  \end{tabular}
  \end{adjustbox}
  \caption{Ablation study results on the multi-scale mechanism used in the framework. Multi.Stru: multi-scale corse-to-fine refinement structure, Multi.Sup: multi-level supervision mechanism. $\uparrow$:the higher, the better.}
  \label{multi}
\end{table}
The importance of the multi-scale mechanism in improving the mIoU performance of the final model is evident. The use of a multi-level supervision approach allows for the deeper levels to capture more general semantic information. Additionally, the multi-scale coarse-to-fine refinement structure facilitates the passing of semantic information from deeper to shallower levels, thereby refining the highest-level feature, which is rich in spatial information. Another key finding from the ablation study is the strong correlation between the multi-scale coarse-to-fine structure and the multi-level supervision mechanism. This structure necessitates supervision signals at each scale to enable the framework to capture relevant semantic information and communicate it to higher levels. Without such supervision signals, this design aspect can impede model convergence and result in suboptimal final performance.

\subsubsection{Ablation on Dynamic Fusion 3D/2D}
\begin{table}[h]
  \centering
  \begin{adjustbox}{width=0.7\columnwidth}
  \begin{tabular}{ccc|c c}
    \toprule
    BEV & SENet2D & SENet3D & mIoU$\uparrow$ & IoU$\uparrow$ \\
    \midrule
    \Checkmark & \Checkmark & \Checkmark & 23.00\% & 43.90\% \\
    \XSolidBrush & \XSolidBrush & \Checkmark & 16.37\% & 43.17\% \\
    \Checkmark & \Checkmark & \XSolidBrush & 19.01\% & 43.10\% \\
    \Checkmark & \XSolidBrush & \Checkmark & 19.94\% & 43.47\% \\ 
    \bottomrule
  \end{tabular}
  \end{adjustbox}
  \caption{Ablation study results on the dynamic fusion 3D/2D module used in the framework. BEV: merged BEV feature,SENet2D: senet block 2D part used to fuse multi-modality BEV features, SENet3D: senet block 3D part used to fuse multi-modality 3D feature volumes. $\uparrow$:the higher, the better.}
  \label{senet}
\end{table}

This study examines the impact of various sub-modules in dynamic fusion 3D/2D on the performance of the final model. After removing the BEV feature, disabling the global-local attention fusion module, the model's mIoU performance decreased by approximately 6.7\% . This observation suggests that the BEV feature plays a crucial role as an excitation signal, facilitating interaction with the 3D feature volume to enable rapid convergence and capture of 3D geometric and semantic information. By removing the SENet3D/2D block in the dynamic fusion module, our model achieves feature fusion solely through feature channel concatenation, devoid of any feature amplification operation. Our model experiences a significant mIoU performance drop as important features remain unamplified.

\section{Conclusion}\label{conclusion}
This study presents OccFusion, a novel framework that integrates surround-view cameras, radars, and lidar to predict 3D semantic occupancy. Our framework employs dynamic fusion 3D/2D modules to consolidate features from diverse modalities, generating a comprehensive 3D volume. The fusion strategies examined in this research encompass Camera+Radar, Camera+Lidar, and Camera+Lidar+Radar combinations. Through a comprehensive evaluation on the nuScenes and SemanticKITTI validation set and subsets focusing on night and rainy scenarios, we observed that even a sparse 3D point cloud from surround-view radars can significantly enhance the vision-centric approach. Moreover, the dense 3D point cloud from lidar further improves 3D occupancy prediction performance. Additionally, we explore how the range of perception influences the performance trends of each sensor fusion strategy across varying perception ranges. Our findings reveal that integrating surround-view radars and lidar can significantly enhance the model's long-distance sensing capabilities and robustness to adverse weather conditions. Our experiments generally exhibit the OccFusion framework's effectiveness while preserving each sensor's distinct strengths.

\bibliographystyle{IEEEtran}
\bibliography{mybibtex}

\begin{thebibliography}{10}
\providecommand{\url}[1]{#1}
\csname url@samestyle\endcsname
\providecommand{\newblock}{\relax}
\providecommand{\bibinfo}[2]{#2}
\providecommand{\BIBentrySTDinterwordspacing}{\spaceskip=0pt\relax}
\providecommand{\BIBentryALTinterwordstretchfactor}{4}
\providecommand{\BIBentryALTinterwordspacing}{\spaceskip=\fontdimen2\font plus
\BIBentryALTinterwordstretchfactor\fontdimen3\font minus \fontdimen4\font\relax}
\providecommand{\BIBforeignlanguage}[2]{{%
\expandafter\ifx\csname l@#1\endcsname\relax
\typeout{** WARNING: IEEEtran.bst: No hyphenation pattern has been}%
\typeout{** loaded for the language `#1'. Using the pattern for}%
\typeout{** the default language instead.}%
\else
\language=\csname l@#1\endcsname
\fi
#2}}
\providecommand{\BIBdecl}{\relax}
\BIBdecl

\bibitem{monoscene}
A.-Q. Cao and R.~de~Charette, ``Monoscene: Monocular 3d semantic scene completion,'' in \emph{Proceedings of the IEEE/CVF Conference on Computer Vision and Pattern Recognition}, 2022, pp. 3991--4001.

\bibitem{bevdet}
J.~Huang, G.~Huang, Z.~Zhu, Y.~Ye, and D.~Du, ``Bevdet: High-performance multi-camera 3d object detection in bird-eye-view,'' \emph{arXiv preprint arXiv:2112.11790}, 2021.

\bibitem{bevformer}
Z.~Li, W.~Wang, H.~Li, E.~Xie, C.~Sima, T.~Lu, Y.~Qiao, and J.~Dai, ``Bevformer: Learning bird’s-eye-view representation from multi-camera images via spatiotemporal transformers,'' in \emph{European conference on computer vision}.\hskip 1em plus 0.5em minus 0.4em\relax Springer, 2022, pp. 1--18.

\bibitem{bevstereo}
Y.~Li, H.~Bao, Z.~Ge, J.~Yang, J.~Sun, and Z.~Li, ``Bevstereo: Enhancing depth estimation in multi-view 3d object detection with temporal stereo,'' in \emph{Proceedings of the AAAI Conference on Artificial Intelligence}, vol.~37, no.~2, 2023, pp. 1486--1494.

\bibitem{tpvformer}
Y.~Huang, W.~Zheng, Y.~Zhang, J.~Zhou, and J.~Lu, ``Tri-perspective view for vision-based 3d semantic occupancy prediction,'' in \emph{Proceedings of the IEEE/CVF Conference on Computer Vision and Pattern Recognition}, 2023, pp. 9223--9232.

\bibitem{occ3d}
X.~Tian, T.~Jiang, L.~Yun, Y.~Wang, Y.~Wang, and H.~Zhao, ``Occ3d: A large-scale 3d occupancy prediction benchmark for autonomous driving,'' \emph{arXiv preprint arXiv:2304.14365}, 2023.

\bibitem{surroundocc}
Y.~Wei, L.~Zhao, W.~Zheng, Z.~Zhu, J.~Zhou, and J.~Lu, ``Surroundocc: Multi-camera 3d occupancy prediction for autonomous driving,'' in \emph{Proceedings of the IEEE/CVF International Conference on Computer Vision}, 2023, pp. 21\,729--21\,740.

\bibitem{radocc}
H.~Zhang, X.~Yan, D.~Bai, J.~Gao, P.~Wang, B.~Liu, S.~Cui, and Z.~Li, ``Radocc: Learning cross-modality occupancy knowledge through rendering assisted distillation,'' \emph{arXiv preprint arXiv:2312.11829}, 2023.

\bibitem{renderocc}
M.~Pan, J.~Liu, R.~Zhang, P.~Huang, X.~Li, L.~Liu, and S.~Zhang, ``Renderocc: Vision-centric 3d occupancy prediction with 2d rendering supervision,'' \emph{arXiv preprint arXiv:2309.09502}, 2023.

\bibitem{bevdet4d}
J.~Huang and G.~Huang, ``Bevdet4d: Exploit temporal cues in multi-camera 3d object detection,'' \emph{arXiv preprint arXiv:2203.17054}, 2022.

\bibitem{panoocc}
Y.~Wang, Y.~Chen, X.~Liao, L.~Fan, and Z.~Zhang, ``Panoocc: Unified occupancy representation for camera-based 3d panoptic segmentation,'' \emph{arXiv preprint arXiv:2306.10013}, 2023.

\bibitem{fbocc}
Z.~Li, Z.~Yu, D.~Austin, M.~Fang, S.~Lan, J.~Kautz, and J.~M. Alvarez, ``Fb-occ: 3d occupancy prediction based on forward-backward view transformation,'' \emph{arXiv preprint arXiv:2307.01492}, 2023.

\bibitem{octreeocc}
Y.~Lu, X.~Zhu, T.~Wang, and Y.~Ma, ``Octreeocc: Efficient and multi-granularity occupancy prediction using octree queries,'' \emph{arXiv preprint arXiv:2312.03774}, 2023.

\bibitem{inversematrixvt3d}
Z.~Ming, J.~S. Berrio, M.~Shan, and S.~Worrall, ``Inversematrixvt3d: An efficient projection matrix-based approach for 3d occupancy prediction,'' \emph{arXiv preprint arXiv:2401.12422}, 2024.

\bibitem{nuscenes}
H.~Caesar, V.~Bankiti, A.~H. Lang, S.~Vora, V.~E. Liong, Q.~Xu, A.~Krishnan, Y.~Pan, G.~Baldan, and O.~Beijbom, ``nuscenes: A multimodal dataset for autonomous driving,'' in \emph{Proceedings of the IEEE/CVF conference on computer vision and pattern recognition}, 2020, pp. 11\,621--11\,631.

\bibitem{LSS}
J.~Philion and S.~Fidler, ``Lift, splat, shoot: Encoding images from arbitrary camera rigs by implicitly unprojecting to 3d,'' in \emph{Computer Vision--ECCV 2020: 16th European Conference, Glasgow, UK, August 23--28, 2020, Proceedings, Part XIV 16}.\hskip 1em plus 0.5em minus 0.4em\relax Springer, 2020, pp. 194--210.

\bibitem{bevdepth}
Y.~Li, Z.~Ge, G.~Yu, J.~Yang, Z.~Wang, Y.~Shi, J.~Sun, and Z.~Li, ``Bevdepth: Acquisition of reliable depth for multi-view 3d object detection,'' in \emph{Proceedings of the AAAI Conference on Artificial Intelligence}, vol.~37, no.~2, 2023, pp. 1477--1485.

\bibitem{bevformerv2}
C.~Yang, Y.~Chen, H.~Tian, C.~Tao, X.~Zhu, Z.~Zhang, G.~Huang, H.~Li, Y.~Qiao, L.~Lu \emph{et~al.}, ``Bevformer v2: Adapting modern image backbones to bird's-eye-view recognition via perspective supervision,'' in \emph{Proceedings of the IEEE/CVF Conference on Computer Vision and Pattern Recognition}, 2023, pp. 17\,830--17\,839.

\bibitem{beverse}
Y.~Zhang, Z.~Zhu, W.~Zheng, J.~Huang, G.~Huang, J.~Zhou, and J.~Lu, ``Beverse: Unified perception and prediction in birds-eye-view for vision-centric autonomous driving,'' \emph{arXiv preprint arXiv:2205.09743}, 2022.

\bibitem{occformer}
Y.~Zhang, Z.~Zhu, and D.~Du, ``Occformer: Dual-path transformer for vision-based 3d semantic occupancy prediction,'' \emph{arXiv preprint arXiv:2304.05316}, 2023.

\bibitem{adabins}
S.~F. Bhat, I.~Alhashim, and P.~Wonka, ``Adabins: Depth estimation using adaptive bins,'' in \emph{Proceedings of the IEEE/CVF Conference on Computer Vision and Pattern Recognition}, 2021, pp. 4009--4018.

\bibitem{AF2S3Net}
R.~Cheng, R.~Razani, E.~Taghavi, E.~Li, and B.~Liu, ``2-s3net: Attentive feature fusion with adaptive feature selection for sparse semantic segmentation network,'' in \emph{Proceedings of the IEEE/CVF conference on computer vision and pattern recognition}, 2021, pp. 12\,547--12\,556.

\bibitem{amvnet}
V.~E. Liong, T.~N.~T. Nguyen, S.~Widjaja, D.~Sharma, and Z.~J. Chong, ``Amvnet: Assertion-based multi-view fusion network for lidar semantic segmentation,'' \emph{arXiv preprint arXiv:2012.04934}, 2020.

\bibitem{polarnet}
Y.~Zhang, Z.~Zhou, P.~David, X.~Yue, Z.~Xi, B.~Gong, and H.~Foroosh, ``Polarnet: An improved grid representation for online lidar point clouds semantic segmentation,'' in \emph{Proceedings of the IEEE/CVF Conference on Computer Vision and Pattern Recognition}, 2020, pp. 9601--9610.

\bibitem{polarstream}
Q.~Chen, S.~Vora, and O.~Beijbom, ``Polarstream: Streaming object detection and segmentation with polar pillars,'' \emph{Advances in Neural Information Processing Systems}, vol.~34, pp. 26\,871--26\,883, 2021.

\bibitem{JS3C-Net}
X.~Yan, J.~Gao, J.~Li, R.~Zhang, Z.~Li, R.~Huang, and S.~Cui, ``Sparse single sweep lidar point cloud segmentation via learning contextual shape priors from scene completion,'' in \emph{Proceedings of the AAAI Conference on Artificial Intelligence}, vol.~35, no.~4, 2021, pp. 3101--3109.

\bibitem{minet}
S.~Li, X.~Chen, Y.~Liu, D.~Dai, C.~Stachniss, and J.~Gall, ``Multi-scale interaction for real-time lidar data segmentation on an embedded platform,'' \emph{IEEE Robotics and Automation Letters}, vol.~7, no.~2, pp. 738--745, 2021.

\bibitem{SPVNAS}
H.~Tang, Z.~Liu, S.~Zhao, Y.~Lin, J.~Lin, H.~Wang, and S.~Han, ``Searching efficient 3d architectures with sparse point-voxel convolution,'' in \emph{European conference on computer vision}.\hskip 1em plus 0.5em minus 0.4em\relax Springer, 2020, pp. 685--702.

\bibitem{cylinder3d}
H.~Zhou, X.~Zhu, X.~Song, Y.~Ma, Z.~Wang, H.~Li, and D.~Lin, ``Cylinder3d: An effective 3d framework for driving-scene lidar semantic segmentation,'' \emph{arXiv preprint arXiv:2008.01550}, 2020.

\bibitem{DRINet++}
M.~Ye, R.~Wan, S.~Xu, T.~Cao, and Q.~Chen, ``Drinet++: Efficient voxel-as-point point cloud segmentation,'' \emph{arXiv preprint arXiv:2111.08318}, 2021.

\bibitem{LidarMultiNet}
D.~Ye, Z.~Zhou, W.~Chen, Y.~Xie, Y.~Wang, P.~Wang, and H.~Foroosh, ``Lidarmultinet: Towards a unified multi-task network for lidar perception,'' in \emph{Proceedings of the AAAI Conference on Artificial Intelligence}, vol.~37, no.~3, 2023, pp. 3231--3240.

\bibitem{voxelnet}
Y.~Zhou and O.~Tuzel, ``Voxelnet: End-to-end learning for point cloud based 3d object detection,'' in \emph{Proceedings of the IEEE conference on computer vision and pattern recognition}, 2018, pp. 4490--4499.

\bibitem{pointpillars}
A.~H. Lang, S.~Vora, H.~Caesar, L.~Zhou, J.~Yang, and O.~Beijbom, ``Pointpillars: Fast encoders for object detection from point clouds,'' in \emph{Proceedings of the IEEE/CVF conference on computer vision and pattern recognition}, 2019, pp. 12\,697--12\,705.

\bibitem{4d-occ}
T.~Khurana, P.~Hu, D.~Held, and D.~Ramanan, ``Point cloud forecasting as a proxy for 4d occupancy forecasting,'' in \emph{IEEE/CVF Conference on Computer Vision and Pattern Recognition (CVPR)}, 2023.

\bibitem{occ4cast}
X.~Liu, M.~Gong, Q.~Fang, H.~Xie, Y.~Li, H.~Zhao, and C.~Feng, ``Lidar-based 4d occupancy completion and forecasting,'' \emph{arXiv preprint arXiv:2310.11239}, 2023.

\bibitem{f-pointnet}
C.~R. Qi, W.~Liu, C.~Wu, H.~Su, and L.~J. Guibas, ``Frustum pointnets for 3d object detection from rgb-d data,'' in \emph{Proceedings of the IEEE conference on computer vision and pattern recognition}, 2018, pp. 918--927.

\bibitem{iPod}
Z.~Yang, Y.~Sun, S.~Liu, X.~Shen, and J.~Jia, ``Ipod: Intensive point-based object detector for point cloud,'' \emph{arXiv preprint arXiv:1812.05276}, 2018.

\bibitem{SIFRNet}
X.~Zhao, Z.~Liu, R.~Hu, and K.~Huang, ``3d object detection using scale invariant and feature reweighting networks,'' in \emph{Proceedings of the AAAI Conference on Artificial Intelligence}, vol.~33, no.~01, 2019, pp. 9267--9274.

\bibitem{pointpainting}
S.~Vora, A.~H. Lang, B.~Helou, and O.~Beijbom, ``Pointpainting: Sequential fusion for 3d object detection,'' in \emph{Proceedings of the IEEE/CVF conference on computer vision and pattern recognition}, 2020, pp. 4604--4612.

\bibitem{MVX-Net}
V.~A. Sindagi, Y.~Zhou, and O.~Tuzel, ``Mvx-net: Multimodal voxelnet for 3d object detection,'' in \emph{2019 International Conference on Robotics and Automation (ICRA)}.\hskip 1em plus 0.5em minus 0.4em\relax IEEE, 2019, pp. 7276--7282.

\bibitem{ContFuse}
M.~Liang, B.~Yang, S.~Wang, and R.~Urtasun, ``Deep continuous fusion for multi-sensor 3d object detection,'' in \emph{Proceedings of the European conference on computer vision (ECCV)}, 2018, pp. 641--656.

\bibitem{bevfusion1}
Z.~Liu, H.~Tang, A.~Amini, X.~Yang, H.~Mao, D.~L. Rus, and S.~Han, ``Bevfusion: Multi-task multi-sensor fusion with unified bird's-eye view representation,'' in \emph{2023 IEEE International Conference on Robotics and Automation (ICRA)}.\hskip 1em plus 0.5em minus 0.4em\relax IEEE, 2023, pp. 2774--2781.

\bibitem{bevfusion2}
T.~Liang, H.~Xie, K.~Yu, Z.~Xia, Z.~Lin, Y.~Wang, T.~Tang, B.~Wang, and Z.~Tang, ``Bevfusion: A simple and robust lidar-camera fusion framework,'' \emph{Advances in Neural Information Processing Systems}, vol.~35, pp. 10\,421--10\,434, 2022.

\bibitem{sparsefusion}
Y.~Xie, C.~Xu, M.-J. Rakotosaona, P.~Rim, F.~Tombari, K.~Keutzer, M.~Tomizuka, and W.~Zhan, ``Sparsefusion: Fusing multi-modal sparse representations for multi-sensor 3d object detection,'' \emph{arXiv preprint arXiv:2304.14340}, 2023.

\bibitem{guizilini2022full}
V.~Guizilini, I.~Vasiljevic, R.~Ambrus, G.~Shakhnarovich, and A.~Gaidon, ``Full surround monodepth from multiple cameras,'' \emph{IEEE Robotics and Automation Letters}, vol.~7, no.~2, pp. 5397--5404, 2022.

\bibitem{EZFusion}
Y.~Li, J.~Deng, Y.~Zhang, J.~Ji, H.~Li, and Y.~Zhang, ``${\mathsf{ezfusion}}$: A close look at the integration of lidar, millimeter-wave radar, and camera for accurate 3d object detection and tracking,'' \emph{IEEE Robotics and Automation Letters}, vol.~7, no.~4, pp. 11\,182--11\,189, 2022.

\bibitem{seeing}
M.~Bijelic, T.~Gruber, F.~Mannan, F.~Kraus, W.~Ritter, K.~Dietmayer, and F.~Heide, ``Seeing through fog without seeing fog: Deep multimodal sensor fusion in unseen adverse weather,'' in \emph{Proceedings of the IEEE/CVF Conference on Computer Vision and Pattern Recognition}, 2020, pp. 11\,682--11\,692.

\bibitem{vehicle}
B.~Major, D.~Fontijne, A.~Ansari, R.~Teja~Sukhavasi, R.~Gowaikar, M.~Hamilton, S.~Lee, S.~Grzechnik, and S.~Subramanian, ``Vehicle detection with automotive radar using deep learning on range-azimuth-doppler tensors,'' in \emph{Proceedings of the IEEE/CVF International Conference on Computer Vision Workshops}, 2019, pp. 0--0.

\bibitem{centerfusion}
R.~Nabati and H.~Qi, ``Centerfusion: Center-based radar and camera fusion for 3d object detection,'' in \emph{Proceedings of the IEEE/CVF Winter Conference on Applications of Computer Vision}, 2021, pp. 1527--1536.

\bibitem{radar}
R.~Yadav, A.~Vierling, and K.~Berns, ``Radar+ rgb fusion for robust object detection in autonomous vehicle,'' in \emph{2020 IEEE International Conference on Image Processing (ICIP)}.\hskip 1em plus 0.5em minus 0.4em\relax IEEE, 2020, pp. 1986--1990.

\bibitem{clrbnn}
R.~Ravindran, M.~J. Santora, and M.~M. Jamali, ``Camera, lidar, and radar sensor fusion based on bayesian neural network (clr-bnn),'' \emph{IEEE Sensors Journal}, vol.~22, no.~7, pp. 6964--6974, 2022.

\bibitem{futr3d}
X.~Chen, T.~Zhang, Y.~Wang, Y.~Wang, and H.~Zhao, ``Futr3d: A unified sensor fusion framework for 3d detection,'' in \emph{Proceedings of the IEEE/CVF Conference on Computer Vision and Pattern Recognition}, 2023, pp. 172--181.

\bibitem{simplebev}
A.~W. Harley, Z.~Fang, J.~Li, R.~Ambrus, and K.~Fragkiadaki, ``Simple-bev: What really matters for multi-sensor bev perception?'' in \emph{2023 IEEE International Conference on Robotics and Automation (ICRA)}.\hskip 1em plus 0.5em minus 0.4em\relax IEEE, 2023, pp. 2759--2765.

\bibitem{resnet}
K.~He, X.~Zhang, S.~Ren, and J.~Sun, ``Deep residual learning for image recognition,'' in \emph{Proceedings of the IEEE conference on computer vision and pattern recognition}, 2016, pp. 770--778.

\bibitem{fpn}
T.-Y. Lin, P.~Doll{\'a}r, R.~Girshick, K.~He, B.~Hariharan, and S.~Belongie, ``Feature pyramid networks for object detection,'' in \emph{Proceedings of the IEEE conference on computer vision and pattern recognition}, 2017, pp. 2117--2125.

\bibitem{senet}
J.~Hu, L.~Shen, and G.~Sun, ``Squeeze-and-excitation networks,'' in \emph{Proceedings of the IEEE conference on computer vision and pattern recognition}, 2018, pp. 7132--7141.

\bibitem{residual}
K.~He, X.~Zhang, S.~Ren, and J.~Sun, ``Deep residual learning for image recognition,'' in \emph{Proceedings of the IEEE conference on computer vision and pattern recognition}, 2016, pp. 770--778.

\bibitem{deformable}
J.~Dai, H.~Qi, Y.~Xiong, Y.~Li, G.~Zhang, H.~Hu, and Y.~Wei, ``Deformable convolutional networks,'' in \emph{Proceedings of the IEEE international conference on computer vision}, 2017, pp. 764--773.

\bibitem{fcos3d}
T.~Wang, X.~Zhu, J.~Pang, and D.~Lin, ``Fcos3d: Fully convolutional one-stage monocular 3d object detection,'' in \emph{Proceedings of the IEEE/CVF International Conference on Computer Vision}, 2021, pp. 913--922.

\bibitem{focal}
T.-Y. Lin, P.~Goyal, R.~Girshick, K.~He, and P.~Doll{\'a}r, ``Focal loss for dense object detection,'' in \emph{Proceedings of the IEEE international conference on computer vision}, 2017, pp. 2980--2988.

\bibitem{lovasz}
M.~Berman, A.~R. Triki, and M.~B. Blaschko, ``The lov{\'a}sz-softmax loss: A tractable surrogate for the optimization of the intersection-over-union measure in neural networks,'' in \emph{Proceedings of the IEEE conference on computer vision and pattern recognition}, 2018, pp. 4413--4421.

\bibitem{atlas}
Z.~Murez, T.~Van~As, J.~Bartolozzi, A.~Sinha, V.~Badrinarayanan, and A.~Rabinovich, ``Atlas: End-to-end 3d scene reconstruction from posed images,'' in \emph{Computer Vision--ECCV 2020: 16th European Conference, Glasgow, UK, August 23--28, 2020, Proceedings, Part VII 16}.\hskip 1em plus 0.5em minus 0.4em\relax Springer, 2020, pp. 414--431.

\bibitem{openoccupancy}
X.~Wang, Z.~Zhu, W.~Xu, Y.~Zhang, Y.~Wei, X.~Chi, Y.~Ye, D.~Du, J.~Lu, and X.~Wang, ``Openoccupancy: A large scale benchmark for surrounding semantic occupancy perception,'' in \emph{Proceedings of the IEEE/CVF International Conference on Computer Vision}, 2023, pp. 17\,850--17\,859.

\bibitem{roldao2020lmscnet}
L.~Rold{\~a}o, R.~de~Charette, and A.~Verroust-Blondet, ``Lmscnet: Lightweight multiscale 3d semantic completion,'' 2020.

\bibitem{li2020anisotropic}
J.~Li, K.~Han, P.~Wang, Y.~Liu, and X.~Yuan, ``Anisotropic convolutional networks for 3d semantic scene completion,'' in \emph{Proceedings of the IEEE/CVF Conference on Computer Vision and Pattern Recognition}, 2020, pp. 3351--3359.

\bibitem{3dsketch}
X.~Chen, K.-Y. Lin, C.~Qian, G.~Zeng, and H.~Li, ``3d sketch-aware semantic scene completion via semi-supervised structure prior,'' in \emph{Proceedings of the IEEE/CVF Conference on Computer Vision and Pattern Recognition}, 2020, pp. 4193--4202.

\bibitem{JS3CNet}
X.~Yan, J.~Gao, J.~Li, R.~Zhang, Z.~Li, R.~Huang, and S.~Cui, ``Sparse single sweep lidar point cloud segmentation via learning contextual shape priors from scene completion,'' in \emph{Proceedings of the AAAI Conference on Artificial Intelligence}, vol.~35, no.~4, 2021, pp. 3101--3109.

\bibitem{li2023voxformer}
Y.~Li, Z.~Yu, C.~Choy, C.~Xiao, J.~M. Alvarez, S.~Fidler, C.~Feng, and A.~Anandkumar, ``Voxformer: Sparse voxel transformer for camera-based 3d semantic scene completion,'' in \emph{Proceedings of the IEEE/CVF conference on computer vision and pattern recognition}, 2023, pp. 9087--9098.

\bibitem{jiang2023symphonize}
H.~Jiang, T.~Cheng, N.~Gao, H.~Zhang, W.~Liu, and X.~Wang, ``Symphonize 3d semantic scene completion with contextual instance queries,'' \emph{arXiv preprint arXiv:2306.15670}, 2023.

\bibitem{udnet}
H.~Zou, X.~Yang, T.~Huang, C.~Zhang, Y.~Liu, W.~Li, F.~Wen, and H.~Zhang, ``Up-to-down network: Fusing multi-scale context for 3d semantic scene completion,'' in \emph{2021 IEEE/RSJ International Conference on Intelligent Robots and Systems (IROS)}.\hskip 1em plus 0.5em minus 0.4em\relax IEEE, 2021, pp. 16--23.

\end{thebibliography}

\end{document}